%% file: main.tex
\documentclass[accepted]{uai2025} 

\input{setup/packages}

\input{setup/commands}

\input{setup/authors}

\begin{document}

\maketitle
\begin{abstract}
Evaluating the performance of clustering models is a challenging task where the outcome depends on the definition of what constitutes a cluster. Due to this design, current existing metrics rarely handle multiple clustering models with diverse cluster definitions, nor do they comply with the integration of constraints when available. In this work, we take inspiration from consensus clustering and assume that a set of clustering models is able to uncover hidden structures in the data. We propose to construct a discriminative ordering through ensemble clustering based on the distance between the connectivity of a clustering model and the consensus matrix. We first validate the proposed method with synthetic scenarios, highlighting that the proposed score ranks the models that best match the consensus first. We then show that this simple ranking score significantly outperforms other scoring methods when comparing sets of different clustering algorithms that are not restricted to a fixed number of clusters and is compatible with clustering constraints.
\end{abstract}

\section{Introduction}
\label{sec:introduction}

\input{sections/1_introduction}

\section{Related works}
\label{sec:related_works}
\input{sections/2_related_works}

\section{The DISCOTEC}
\label{sec:method}
\input{sections/3_method}

\section{Experiments}
\label{sec:experiment}
\input{sections/4_experiment}

\section{Conclusion}
\label{sec:conclusion}
\input{sections/5_conclusion}

\begin{acknowledgements}
This research is financially supported by the Swedish Research Council via projects 2020-04122 and 2024-05011, the Knut and Alice Wallenberg Foundation via project KAW 2020.0033 and the Wallenberg AI, Autonomous Systems and Software Program (WASP), and the Excellence Center at Linköping--Lund in Information Technology (ELLIIT).
\end{acknowledgements}

\bibliography{bib}

\clearpage
\newpage
\onecolumn

\title{Discriminative ordering through ensemble consensus\\(Supplementary Material)}
\maketitle

\appendix

\section{Supplementary details for experiments}
\label{app:benchmark_details}
\input{appendix/benchmark_details}

\section{Scaling with the number of models under synthetic scenarios}
\label{app:scaling}
\input{appendix/scaling_experiment}

\section{Consensus matrix visualisation}
\label{app:consensus_visualisation}
\input{appendix/consensus_visualisation}

\section{Extended benchmark results}
\label{app:extended_benchmark}
\input{appendix/extended_benchmark}


\end{document}

%% file: setup/packages.tex
\usepackage{natbib} 
    \setcitestyle{round, authoryear}

\usepackage{microtype}
\usepackage{graphicx}
\usepackage{booktabs} 

\usepackage{multirow}

\usepackage{hyperref}

\usepackage{algpseudocode}
\usepackage{algorithm}

\usepackage{amsmath}
\usepackage{amssymb}
\usepackage{mathtools}
\usepackage{amsthm}
\usepackage{stmaryrd}
\usepackage{bbm}

\usepackage{subfig}

\usepackage{pgfplots}
\pgfplotsset{compat=1.6}

\usepackage[capitalize,noabbrev]{cleveref}

\theoremstyle{plain}

\theoremstyle{definition}

\theoremstyle{remark}

%% file: setup/commands.tex
\newcommand{\std}[1]{\textsubscript{#1}}
\newcommand{\ie}{\textit{i.e.} }
\newcommand{\eg}{\textit{e.g.} }

%% file: setup/authors.tex
\title{Discriminative ordering through ensemble consensus}

\author[1]{\href{mailto:<louis.ohl@liu.se>?Subject=Your UAI 2025 paper}{Louis~Ohl}}
\author[1]{Fredrik~Lindsten}
\affil[1]{%
    Department of Computer and Information Science\\
    Linköping University\\
}

%% file: sections/1_introduction.tex
Clustering is an essential task in data analysis where one seeks to partition the observations of a dataset into $K$ clusters. Due to its ill-posed nature, the design of a clustering algorithm requires hypotheses about what defines good clusters. Different hypotheses may lead to different clusters. In other words, cluster definition and methodology must be adapted to the context in which they are applied~\citep{hennig_what_2015}.

Evaluating the quality of a clustering model is a complex problem that requires appropriate metrics. In an experimental setting, synthetic data can be generated according to hypotheses about the definition of clusters, allowing verification that a clustering algorithm recovers the expected partition. This verification can be done using \emph{external} metrics such as the (unsupervised) accuracy, the normalised mutual information (NMI), or the adjusted Rand index (ARI, \citealp{hubert_comparing_1985}). Conversely, in an exploratory context, \ie when no labels are available, we rely on \emph{internal} metrics that depend solely on the data observations and the model predictions, \eg the variance-ratio criterion~\citep{calinski_dendrite_1974},  the silhouette score~\citep{rousseeuw_silhouettes_1987}, or the integrated complete likelihood~\citep{biernacki_assessing_2000}. Internal metrics are often built with a specific view on clustering hypotheses, and therefore must be used with algorithms that match those hypotheses.

Despite the large number of clustering metrics~\citep{desgraupes_clustering_2013, charrad_nbclust_2014}, there are few metrics that are suitable for comparing clustering models with different clustering hypotheses. In addition, and to the best of our knowledge, there is no clustering metric that can integrate constraints when targets are partially observed.

To compare different clustering algorithms independently of their clustering hypotheses, we take inspiration from consensus clustering~\citep{strehl_cluster_2002} and rank clustering algorithms according to their proximity to a consensus matrix. Our underlying hypothesis is that a diverse set of clustering algorithms will shed light on clusters whose observations are more frequently connected.
Our contributions are:

\begin{itemize}
    \item The proposal of a simple-to-compute and fast score for ranking clustering algorithms based on consensus clustering that is compatible with pairwise constraints regularisations.
    \item The first exploration of clustering ensembles as a mean of performing model selection, both for our metric and some baselines.
    \item An extensive benchmark including synthetic and real data for several internal metrics showing strong performances in favour of our metric.
\end{itemize}

%% file: sections/2_related_works.tex
Evaluating the quality of a clustering model is a challenging task. In fact, the absence of a formal definition of what a cluster is leads to the absence of a definition of what quality is, and finding an objective measure of quality that allows comparison of different algorithms is challenging~\citep{boley_partitioning-based_1999}. For example, \citet{han_10_2012} define quality as ``\textit{[s]ome methods [that] measure how well the clusters fit the data set, while others measure how well the clusters match the ground truth}''. These two categories can be referred to as internal metrics and external metrics.

\subsection{External indices}

In the presence of ground truth, \ie targets, we can evaluate clustering models using external indices. Common evaluation metrics include the (unsupervised) accuracy, the NMI~\citep{strehl_cluster_2002} or the ARI~\citep{hubert_comparing_1985}. It is worth noting that NMI and ARI are preferable to unsupervised accuracy when the number of clusters in a model differs from the number of clusters in the targets, which is unknown in practice.

When we partially observe the targets, we can still use them as constraints, as in semi-supervised clustering~\citep{bair_semi-supervised_2013, cai_review_2023}. Two types of constraints can be distinguished: labels and must-link or cannot-link constraints. The former explicitly assign samples to a cluster, whereas the latter only indicate whether samples should be together or in different clusters, regardless of the cluster membership. Labels imply must-link and cannot-link constraints, but the reverse is not true. For example, we can evaluate the model using the pairwise recall, precision and F-measure~\citep{basu_active_2004}, or the constrained Rand index (CRI, \citealp{klein_instance-level_2002}). For both measures, the evaluation is restricted to the set of samples (or set of sample pairs) that are not affected by constraints. These metrics are external and require ground truth. Consequently, they cannot be used if we do not have access to labels other than those used to constrain the clustering algorithm. In the absence of such information, the approximate measure of informativeness~\citep{davidson_measuring_2006} could be preferred: it is simply the average number of constraints not satisfied by a clustering algorithm.

\subsection{Internal indices}

We can distinguish two types of internal indices: those that integrate the clustering hypotheses from the model, and those that carry their own hypotheses about the definition of what makes good clusters.

If a model defines a tractable likelihood, we can use this value to reflect the fit between the model and the data. For example, the Akaike information criterion (AIC, \citealp{akaike_new_1974, akaike_information_1998}) penalises the likelihood by the complexity of the model, expressed in terms of the number of free parameters. The Bayesian information criterion (BIC; \citealp{schwarz_estimating_1978}) weights this parameter penalty by the logarithm of the number of training samples. The integrated complete likelihood (ICL, \citealp{biernacki_assessing_2000}) extends the BIC in model-based clustering by distinguishing between model components in model-based clustering and their correspondence to clusters using an entropy penalty term.

If a model does not define a tractable likelihood, we may not have access to a fitness measure from the model and have to construct it post hoc. This is notably the case for discriminative clustering models, \eg KMeans or DBSCAN~\citep{ester_density-based_1996}. In this sense, the most well-known internal metric is the within-group sum of squares (WGSS, \citealp{edwards_method_1965}), also known as the KMeans score. This score is efficient for clusters that are assumed to be concentrated around a centroid. However, ensuring that samples are concentrated around a centroid is not enough; a clear separation between clusters is also a desirable property. From this desire come criteria such as  the variance-ratio criterion~\citep{calinski_dendrite_1974},  the Dunn index~\citep{dunn_well-separated_1974} and its generalisations~\citep{bezdek_new_1998}, which include the Davies-Bouldin index~\citep{davies_cluster_1979}, the Silhouette score~\citep{rousseeuw_silhouettes_1987} or the PBM index~\citep{pakhira_validity_2004}. Internal metrics comparing the coherence of pairwise clustering have also been proposed. For instance, the Gamma index~\citep{baker_measuring_1975} and the G+ index~\citep{rohlf_methods_1974} are based on the notion of discordant and concordant pairs. A pair of samples from a similar cluster is concordant with another pair of samples from different clusters when their distance is shorter than the second pair. The two pairs are discordant if the distance is greater for dissimilar clusters than for similar clusters. To alleviate the requirement on the choice of distances, some of these scores were adapted for connectivity matrices~\citep{saha_connectivity_2012} derived from relative neighbourhood graphs~\citep{toussaint_relative_1980}.

\subsection{Ranking models in consensus clustering}

If we restrict the goal of a clustering metric to comparing models, then the most relevant property is the ability of a metric to \emph{rank} algorithms well. In a ranking context, we necessarily have several models: this allows us to use ensemble methods. Consensus clustering is an unsupervised ensemble clustering method that stems from classification ensembles~\citep{strehl_cluster_2002}. The goal is to use several clustering algorithms, called base clusterings, and to combine their results into a single final clustering using a consensus function. Combining the results thus increases the quality of the clustering, in the sense of an evaluation using external labels.

Several works then developed some filtering criteria on the base clusterings to improve the quality of consensus clustering. This field, sometimes called \emph{ensemble clustering selection}~\citep{golalipour_clustering_2021}, focuses on selecting a subset of base clusterings based on the belief that some of the base models hinder the global quality and should be discarded. The goal is to keep clusterings of quality while maintaining some diversity~\citep{kuncheva_using_2004, hadjitodorov_moderate_2006, fern_cluster_2008}. This selection can be done by keeping the base clusterings that are closest to the consensus result~\citep{hong_resampling-based_2009, azimi_adaptive_2009, jia_bagging-based_2011}. Although this introduces ordering between models, this ordering is non-deterministic as it relies on the outcome of the consensus which can be stochastic. Selection can also be achieved by solving a K-vertex subgraph problem on a graph, where edges are the similarities between pairs of base clusterings~\citep{fern_cluster_2008, yang_cluster_2017}. However, such an approach does not introduce an order between models.

In some cases, this selection is made thanks to a ranking. Often, this ranking is done by interpolating between the quality and the diversity of each clustering algorithm~\citep{fern_cluster_2008, naldi_cluster_2013, wang_two-level-oriented_2018}. The ranking then depends simultaneously on the definition of what is the quality of a base clustering and what diversity represents, in an internal metric sense, and on the interpolation coefficient. Thus, metrics for ranking clustering algorithms are not new, \emph{but their purpose is different}. Therefore, and to the best of our knowledge, ranking clustering algorithms through consensus has never been used as a metric for selecting clustering models.

%% file: sections/3_method.tex
We seek to build a score for ranking clustering algorithms that simultaneously take into account the results of all algorithms and comply with must-link and cannot-link constraints. We start with the general unconstrained score, then detail how it can be simplified thanks to consensus binarising and finish with the addition of constraints.

\subsection{The unconstrained score}

We assume that we have a set of $T\geq 3$ clustering models and a dataset of $n$ unlabelled samples $\mathcal{D} = \{x_i\}_{i=1}^n$. Each clustering model $t$ defines a partition of this dataset into $K^t$ clusters: $\pi^t \in \{1, \ldots, K^t\}^n$. Note that we only consider hard clusterings here, so that our method is compatible with any clustering algorithm, since soft clusterings can always be converted to hard ones.

We construct for each partition its respective connectivity matrix. Its entries are binary values indicating whether two samples were in the same cluster:
\begin{equation}
    \pmb{A}^t = \left[ \mathbbm{1}[\pi^t_i = \pi^t_j]\right].
\end{equation}
We can then build the consensus matrix \citep{monti_consensus_2003} by averaging all connectivity matrices:
\begin{equation}
    \pmb{C} = T^{-1} \sum_{t=1}^T \pmb{A}^t.
\end{equation}
The entries of the consensus matrix can be interpreted as parameters of Bernoulli distributions: they describe the probability that two samples end up in the same cluster according to the ensemble of models. The more often a pair of observations end up in the same cluster, the higher their consensus value. Consequently, we would like to identify a clustering that respects this trend. Conversely, when the consensus value is close to zero, we would like to select a clustering that did not link the two observations.

To order the clustering algorithms, we propose to measure the distance between their respective connectivity matrix and the consensus matrix. The smaller the distance, the better. We expect that the model with the lowest distance corresponds to a partition that best matches the consensus established by the ensemble. For an arbitrary distance or divergence $D$, \eg the total variation distance or the KL divergence:
\begin{equation}
    \mathcal{S}(\pi^t) = \sum_{i,j} D(\pmb{A}^t_{ij} \| \pmb{C}_{ij}).
\end{equation}
Note that some combinations of inputs are impossible when computing $D$. We cannot have 0 (resp. 1) for connectivity $\pmb{A}_{ij}$ and 1 (resp. 0) for the consensus $\pmb{C}_{ij}$ because the consensus is an average. When both values are 0, or 1, the distance is necessarily 0. Therefore, the only distances we compute correspond to the cases where $\pmb{C}_{ij} \in ]0,1[$. We summarise three examples of distances for this case, which we will use in experiments, in Table~\ref{tab:explicit_distances}.

\input{figures/table_explicit_distances}

It is possible that this score favours solutions with too few or too many clusters. In fact, when clustering models tend to connect most of the samples together through large clusters, the ranking would favour solutions with few clusters because they minimise the number of terms $D(\mathcal{B}(0)\|\mathcal{B}(\pmb{C}_{ij}))$, which incur a large penalty. Conversely, when most clustering models have a large number of clusters, the consensus matrix may become sparse or filled with very low values and the score would favour solutions with many clusters because they minimise the number of terms $D(\mathcal{B}(1)\|\mathcal{B}(\pmb{C}_{ij}))$. When the number of clusters varies from both extremes in the pool of clustering models, then the behaviour of the score would be in favour of solutions with many clusters because the consensus matrix gets low values. 

\input{figures/algorithm_discotec}

In order to alleviate the limitation of having only high values or only low values in the consensus matrix, we propose to binarise it with respect to its mean:
\begin{equation}
\pmb{Q} = \left[\mathbbm{1}\left[\pmb{C}_{ij} \geq n^{-2} \sum_{i^\prime j^\prime} \pmb{C}_{i^\prime j^\prime}\right]\right].
\end{equation}
In this variant, we measure only the absolute differences between zeros and ones from both the connectivity and the consensus matrices. While this binarised consensus matrix is not compatible with the original perspective of statistical distances between two matrices, it can be interpreted as the ratio of mismatching connectivities between observations: the lower the better.

\subsection{Adding regularisations}

An important feature of the proposed score is its compatibility with the approximate measure of informativeness~\citep{davidson_measuring_2006}, \ie the average number of violated constraints. Given a set of $n_\text{ML}$ must-link constraints $\mathcal{C}_{n_\text{ML}} = \{(a_i,b_i)\}_{i=1}^{n_\text{ML}}$, and a set of $n_\text{CL}$ cannot-link constraints $\mathcal{C}_{n_\text{CL}} = \{(a_i,b_i)\}_{i=1}^{n_\text{CL}}$, this regularisation is:
\begin{equation}
\mathcal{R}(\pi^t) = \frac{\sum_{(a,b) \in \mathcal{C}_\text{ML}} D(\pmb{A}_{ab}^t\| 1) +\sum_{(a,b) \in \mathcal{C}_\text{CL}} D(\pmb{A}_{ab}^t\| 0)}{n_\text{ML}+n_\text{CL}}.
\end{equation}
Both the regularisation and our score are contained in [0,1] and correspond to the sum of distances between a connectivity and a target value. Thus, both measures are compatible according to dimensional analysis.

We summarise the binarised version of the discriminative ordering through ensemble consensus (DISCOTEC) in Algorithm~\ref{alg:binary_discotec_algorithm}. We evaluate the computational complexity of this algorithm to $\mathcal{O}(T(n^2+n_\text{ML}+n_\text{CL}))$ for $T$ models and $n$ observations.

We may note that the DISCOTEC scales linearly with the number of models. In comparison, the average NMI (ANMI, \citealp{strehl_cluster_2002}) and the average ARI, which were used for clustering ensemble selection~\citep{fern_cluster_2008}, scale quadratically. These metrics consist in the average of a score between a partition and all other partitions, \eg for ANMI:
\begin{equation}
    \text{ANMI}(\pi^t) = \frac{1}{T-1} \sum_{t^\prime\neq t} \text{NMI}(\pi^t, \pi^{t^\prime}).
\end{equation}
Consequently, evaluating the AARI or ANMI requires $T(T-1)/2$ pairwise computations, which becomes expensive when the number of models is large.

%% file: figures/table_explicit_distances.tex
\begin{table}[t]
    \centering
    \caption{Examples of formula for the distance $D$ between connectiviy and consensus with different statistical distances.}
    \label{tab:explicit_distances}
    \begin{tabular}{r c c}
         \toprule
         &$D(\mathcal{B}(0) \| \mathcal{B}(\pmb{C}_{ij}))$&$D(\mathcal{B}(1) \| \mathcal{B}(\pmb{C}_{ij}))$\\
         \midrule
         KL&$-\log (1-\pmb{C}_{ij})$&$-\log \pmb{C}_{ij}$\\
         TV&$\pmb{C}_{ij}$&$1-\pmb{C}_{ij}$\\
         H\textsuperscript{2}&$1-\sqrt{1-\pmb{C}_{ij}}$&$1-\sqrt{\pmb{C}_{ij}}$\\
         \bottomrule
    \end{tabular}
\end{table}

%% file: figures/algorithm_discotec.tex
\begin{algorithm}[t]
\centering
\caption{The binarised DISCOTEC}
\label{alg:binary_discotec_algorithm}
\begin{algorithmic}
\Require A set of partitions $\pi^t \in \{1, \ldots, K^t\}^n; t\in \{1, \ldots,T\}$.
\Require Must-link constraints $\mathcal{C}_\text{ML} = \{(a_i, b_i)\}_{i=1}^{n_\text{ML}}$
\Require Cannot-link constraints $\mathcal{C}_\text{CL} = \{(a_i, b_i)\}_{i=1}^{n_\text{CL}}$
\For{$t\in \{1, \ldots, T\}$}
\State $\pmb{A}^t \gets \left[\mathbbm{1}[\pi^t_i=\pi^t_j]\right]$ \Comment{Connectivity matrices}
\EndFor
\State $\pmb{C} \gets T^{-1} \sum_{t=1}^T \pmb{A}^t$ \Comment{Consensus matrix}
\State $\mu \gets n^{-2}\sum_{i,j}^{n} \pmb{C}_{ij}$
\State $\pmb{Q} \gets \left[\mathbbm{1}[\pmb{C}_{ij} \geq \mu]\right]$ \Comment{Binarise the consensus}
\For{$t\in \{1, \ldots, T\}$}
\State $\mathcal{S}^t \gets n^{-2}\sum_{i,j}^{n}\left|\pmb{Q}_{ij}-\pmb{A}_{ij}^t\right|$ \Comment{Score of model $t$}
\State $\mathcal{R}^t \gets 0$ \Comment{Regularisation by constraints}
\For{$(i,j) \in \mathcal{C}_\text{ML}$}
\State $\mathcal{R}^t \gets \mathcal{R}^t + (1-\pmb{A}_{ij}^t)$
\EndFor
\For{$(i,j) \in \mathcal{C}_\text{CL}$}
\State $\mathcal{R}^t \gets \mathcal{R}^t + \pmb{A}_{ij}^t$
\EndFor
\State $\mathcal{S}^t \gets \mathcal{S}^t + \frac{\mathcal{R}^t}{n_\text{ML}+n_\text{CL}}$ \Comment{Regularised DISCOTEC}
\EndFor
\end{algorithmic}
\end{algorithm}

%% file: sections/4_experiment.tex
For our experiments, we incrementally moved from synthetic partitions to datasets and constraints integration. We first show that the DISCOTEC and other ensemble baselines perform on par on synthetic cases. We then introduce datasets and test both clustering algorithms with a fixed number of clusters or an unrestricted number of clusters. We highlight that the DISCOTEC has strong performances for the latter. Finally, we show that constraints can enhance the performance of DISCOTEC on real datasets, even with few constraints.

\subsection{General protocol}

To evaluate the DISCOTEC, we have borrowed the methodology of \citet[section 4]{vendramin_relative_2010}. We first select a pool of $N_\mathcal{D}$ datasets, and for each dataset we apply $T$ clustering algorithms. We then evaluate the correlation between an internal metric of interest and an external metric that describes how well a model matches some targets. A higher correlation value indicates that the ranking proposed by the internal metric is efficient in identifying the most relevant clustering. We report the average correlations over the $N_\mathcal{D}$ datasets. Note that we have negated all scores that should be minimised, so a positive correlation means good performance. We chose to show the Kendall's tau correlation~\citep{kendall_treatment_1945} in the paper because it measures how well two rankings compare. For extended results, including the Pearson correlation as originally proposed by \citet{vendramin_relative_2010}, see Appendix~\ref{app:extended_benchmark}.

We distinguish two types of baselines: internal metrics that are also based on ensemble clustering and internal metrics that evaluate models individually using distances between observations. For the former, we use the ANMI and AARI, and emphasise that this is the first time the ranking properties of these metrics is studied. For the latter, we used clustering metrics available in the permetrics Python library~\citep{thieu_permetrics_2024} and implemented ourselves some from the clusterCrit package~\citep{desgraupes_clustering_2013}. For the sake of clarity, we have restricted all figures and tables to the ensemble metrics and the top-performing distance-based metrics where relevant. Extended tables with the 20 baselines can be found in Appendix~\ref{app:extended_benchmark}. Code can be found at: \url{https://github.com/oshillou/Discotec}.

\subsection{Synthetic partitions}

\input{figures/fig_synthetic_partitions_correlations}
\input{figures/fig_synthetic_pivot}

We started by evaluating the DISCOTEC with synthetic partitions, which allowed us to control the difficulty of the consensus. We started by generating a ground truth of $n$ observations and $K$ clusters, then generated $T$ different partitions trying to imitate the ground truth with some controlled accuracy. To that end, we sample for each observation a conservation indicator according to some probability $\rho$. If the observation is conserved, it keeps the same cluster as the ground truth. Otherwise, it is assigned to a different cluster than the ground truth.

We tested two synthetic scenarios: one with a uniform distribution of accuracies to the ground truth and one with unbalanced accuracies. For the first scenario, we uniformly sample a conservation threshold $\rho^t \in [0.1, \rho_\text{max}]$ for each model. This ensures that the models have a minimum accuracy of 10\%, and an average maximum accuracy of $\rho_\text{max}$. For the second scenario, we first sample two partitions called \emph{hubs}: one with $\rho=0.2$ and one with $\rho=0.9$. Then, we sample a fraction $\alpha T$ of the models with an accuracy in the range $[0.2, 0.9]$ to the first hub, and the remaining $(1-\alpha)T$ models with identical accuracy range to the second hub.

Since both of these scenarios do not have any underlying data samples on which we can measure distances, we can only evaluate the AARI, the ANMI and the DISCOTEC. We ran both scenarios with $n=200$ samples and $K=10$ clusters. In the first scenario, we varied $T$ from 5 to 50 models. In the second scenario, we fixed $T=50$. Each simulation was repeated 50 times. The results of the first scenario are shown in Figure~\ref{fig:synthetic_partitions_correlation} and of the second scenario in Figure~\ref{fig:synthetic_partitions_pivot}. For the sake of readability, we only report the DISCOTEC with KL divergence and with binarised consensus in the figures, as the rankings using total variation distance and squared Hellinger distance followed the KL curve perfectly.

From the first scenario, we observe that increasing the maximum possible accuracy with $\rho_\text{max}$ increases the correlation of the ranking. Indeed, when the maximum accuracy is low, most of the synthetic partitions tend to disagree with each other, resulting in a noisy consensus. Consequently, no pattern can emerge from the consensus matrix, and the DISCOTEC fails to correctly identify the correct clusterings. In contrast, if the maximum accuracy is high, a pattern can be seen in the consensus matrix, and the ranking can be coherent with this pattern. For completeness, we have included examples of such matrices in Appendix~\ref{app:consensus_visualisation}. We can note in Figure~\ref{fig:synthetic_partitions_correlation} that the number of models is crucial to improve the performance of both baselines and DISCOTEC. Indeed, the correlation between the ARI of the partitions on the targets and the ranking of each method increases, and its standard deviation decreases from 5 to 50 models. This effect is even stronger for the binarised DISCOTEC. We further discuss and experiment with scaling within this scenario in Appendix~\ref{app:scaling}.

The success of the first scenario is due to the uniform distribution in terms of accuracy of all sampled models, but does not transfer to the second scenario. The second scenario highlights that both the DISCOTEC and the baselines are attracted to dominant hubs in terms of clustering solutions. Indeed,  we can see in Figure~\ref{fig:synthetic_partitions_pivot} that when the partitions are close to a solution with high accuracy, \ie $\alpha\approx 0$, then the ranking has a high correlation with the ARI. Conversely, increasing the number of models that are similar to a poor solution with very low accuracy, \ie $\alpha=1$, decreases the correlation for the same reason of noisy patterns as described above.

In summary, we have shown with these synthetic scenarios that ranking according to the relationships between models, both in baselines and the DISCOTEC, depends on two main factors: (i) the number of models and (ii) the distribution of the clustering ARIs. The number of models should preferably be large enough. However, too large a number of models is detrimental to the AARI and ANMI, which scale quadratically while the DISCOTEC scales linearly. Regarding the distribution of the clustering ARIs, we can expect better performance if it is more concentrated on solutions that are close enough to the ground truth and uniform. In other words: the diversity of base clusterings matters, in the sense of different cluster definitions.

\subsection{Synthetic and real datasets clustering}

To simulate more complex distributions of ARI with respect to targets, we now turn to different combinations of clustering models and datasets. We considered two different categories of datasets for our experiment: the fundamental clustering problem suite (FCPS, \citealp{thrun_fundamental_2021}) and real datasets from the UCI repository, summarised in Appendix~\ref{app:benchmark_details}. The FCPS consists of different simulated datasets in two or three dimensions, so that the definition of clusters is consensual to the naked eye. In contrast, the UCI datasets are intended for classification, which means that the classes and their number may not reflect the clusters and their number. Therefore, we must be careful in our interpretation of the ARI depending on the category of the datasets.

\subsubsection{Restriction to a fixed number of clusters}

\input{figures/table_benchmark_fixed_kendall_restricted}
\input{figures/table_benchmark_fixed_regretari_restricted}

Similarly to the synthetic scenarios, we restrict our experiments to clustering models that must find as many clusters as the number of clusters (resp. classes) indicated by the targets of the FCPS (resp. UCI) datasets. We try two different algorithms: KMeans and agglomerative clustering. We run KMeans 50 times. Since agglomerative clustering deterministically produces the same clustering, we vary its parameters using single, average, complete, and Ward linkage, and also Euclidean or Manhattan distance. This results in 7 models, because the Manhattan distance and the Ward linkage are incompatible.

Following the general protocol, we report the correlation in Table~\ref{tab:benchmark_fixed_kendall_restricted}. Since the average correlation can be high due to some lucky runs, we extend our results by also reporting the regret score on the ARI of the top-ranked model for all methods. We define the regret score as the difference between the best performance of all methods and the performance of one method, which we average over all datasets. A lower regret score is better, and a regret score of 0 indicates that the method always had the best performance. We report the regret scores in Table~\ref{tab:benchmark_fixed_regretari_restricted}. Regret scores on the correlation can be found in Appendix~\ref{app:extended_benchmark}.

These results complement the observations made previously in our second synthetic scenario. Indeed, we can see in Table~\ref{tab:benchmark_fixed_regretari_restricted} that the clusterings proposed by the agglomerative algorithms are more diverse than for KMeans since the ARI regret is up to 38\% behind for some scores. This diversity leads to higher correlations compared to KMeans algorithms in Table~\ref{tab:benchmark_fixed_kendall_restricted}. In contrast, the KMeans algorithms were attracted to specific clusterings that had a low ARI with respect to the targets for some datasets, leading to lower correlations. Furthermore, the lack of diversity between the base clusterings  and the regret on the selected model ARI is similar for all scores.

Among the compared baselines, we do not distinguish any score that offers a better ranking than any other for this experiment. We only mention both the silhouette index (SI, \citealp{rousseeuw_silhouettes_1987}) as an example and the strong success of the Calinski-Harabasz index (CHI, \citealp{calinski_dendrite_1974}) for the UCI datasets with agglomerative clustering, highlighted by a high correlation in Table~\ref{tab:benchmark_fixed_kendall_restricted} and a regret score of 0 on the selected model ARI in Table~\ref{tab:benchmark_fixed_regretari_restricted}.

\subsubsection{Unrestricted pool of clustering models}
\input{figures/table_benchmark_mixed_kendall_restricted}
\input{figures/table_benchmark_mixed_regretari_restricted}

We now extend the previous experiments by proposing a more diverse pool of clustering algorithms. We run KMeans clustering with $K$ varying from 2 to 20 for each dataset, 5 times per value of $K$. We run agglomerative clustering with the same linkage parameters as before with Euclidean distances for 2 to 20 clusters. We then add DBSCAN models with parameter epsilon varying from the 1\% quantile of the Euclidean distances of the dataset to the 25\% quantile. We discard degenerate clusterings. Finally, we also evaluate the performance of the ranking methods when we merge all the models.

We observe in Table~\ref{tab:benchmark_mixed_kendall_restricted} that the average correlation is the highest for the DISCOTEC with binarised consensus matrix. Moreover, the ARI regret of the selected model is also the lowest in Table~\ref{tab:benchmark_mixed_regretari_restricted}, which reveals better selection. In contrast, the DISCOTEC using the KL divergence did not perform better than the AARI baselines.

The performance of the DISCOTEC with KL divergence suffered precisely from the overclustering bias that motivated the introduction of the binarised consensus. The high proportion of models with a large number of clusters contributed to lowering the values of the consensus matrix, bringing them all close to 0. The KL ranking consequently favoured models with the largest number of clusters because they are less penalised when connecting as few observations as possible. 

Finally, there is a notable difference between the FPCS and UCI datasets. For the former, we have the certainty that at least one of the proposed algorithms will achieve an ARI of 1 when merged together,  because it matches the empirical definition of the clusters in a dataset. In contrast, the latter does not guarantee that the targets reflect clusters. Therefore, it is likely that the set of clustering algorithms will point to different clusters than the targets, sometimes with a different number of clusters compared to the number of classes. This accounts for the lower correlation values for the UCI datasets compared to the FCPS datasets in Table~\ref{tab:benchmark_mixed_kendall_restricted}.

\subsection{Constraint integration}
\input{figures/fig_regularisation}

To make the most sense of the correlation between targets and clusters in the UCI datasets, we add constraints to the ranking. Thus, we simultaneously look for a model that captures clusters that correspond to what most models find, while also respecting the classes as much as possible.

We measured constraint satisfaction using the approximate measure of informativeness $\mathcal{R}$, and added it to both the DISCOTEC and the AARI/ANMI baselines. We chose not to add it to distance-based metrics because they do not correspond to the approximate measure of informativeness in terms of dimensional analysis. Moreover, the constraint regularisation is bounded in [0,1], whereas some other metrics are unbounded.

To assess the benefit of constraints, we report for each dataset the initial unconstrained correlation between targets and rankings using the same models from the previous experiment, and report the correlation after adding constraints. For each dataset, we randomly selected $n$ observations and generated all must-link and cannot-link constraints they implied using the targets, and then evaluated the correlation of the regularised rankings. We repeated the constraint addition 50 times. Correlations before and after constraint addition can be found in Figure~\ref{fig:benchmark_correlation_gain}.

We observe that the addition of constraints is rarely detrimental to the correlation, as highlighted by the standard deviation decreasing from 0 constrained observations to 5 constrained observations. Moreover, increasing the number of constraints increases the correlation of the ranking with the targets. However, this increase is more substantial when introducing the first few constraints and tends to flatten afterwards. Nonetheless, we may note that 50 constrained observations is relatively small for some UCI datasets , \eg Segmentation with 2310 observations.

%% file: figures/fig_synthetic_partitions_correlations.tex
\begin{figure*}[!th]
    \centering
    \subfloat[5 models]{
        \includegraphics[width=0.32\linewidth]{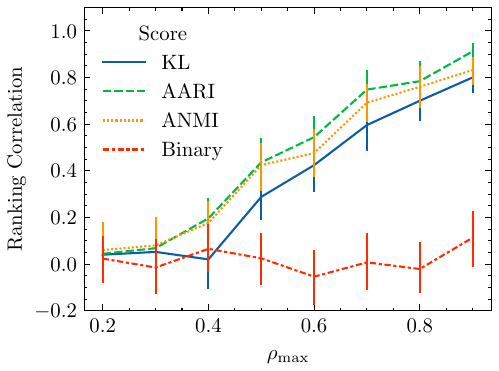}
    }
    \subfloat[25 models]{
        \includegraphics[width=0.32\linewidth]{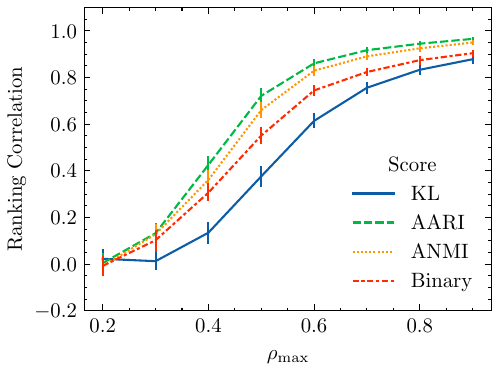}
    }
    \subfloat[50 models]{
        \includegraphics[width=0.32\linewidth]{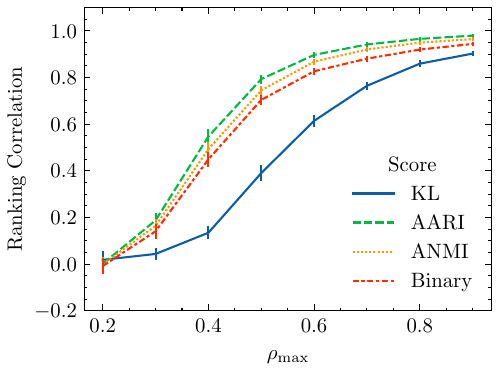}
    }
    \caption{Evolution of the average Kendall's tau ($\uparrow$) correlation between ranking metrics and ARI with targets of synthetic partition when the label preservation rate $\rho_\text{max}$ increases.}
    \label{fig:synthetic_partitions_correlation}
\end{figure*}

%% file: figures/fig_synthetic_pivot.tex
\begin{figure}[!th]
    \centering
        \includegraphics[width=0.8\linewidth]{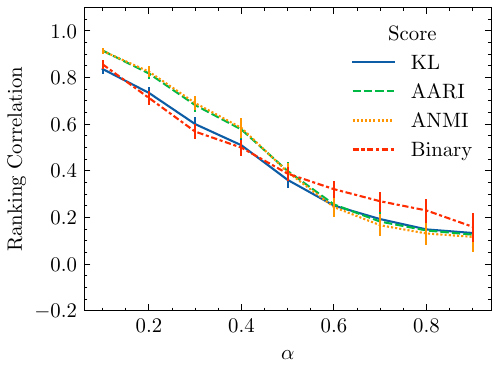}
    \caption{Evolution of Kendall's tau correlation ($\uparrow$) of the selected models per ranking method as the interpolation $\alpha$ varies between highly accurate models ($\alpha=0$) and non-accurate models ($\alpha=1$).}
    \label{fig:synthetic_partitions_pivot}
\end{figure}

%% file: figures/table_benchmark_fixed_kendall_restricted.tex
\begin{table}[!th]
    \centering
    \caption{Average Kendall tau correlation\std{std} ($\uparrow$) of the ranking metrics when the base clusterings are restricted to as many clusters as targets. KL and Binary respectively stand for the DISCOTEC with KL divergence, and consensus binarisation.}
    \label{tab:benchmark_fixed_kendall_restricted}
    \begin{tabular}{lllll}
        \toprule
        &\multicolumn{2}{c}{FCPS}&\multicolumn{2}{c}{UCI}\\
        \cmidrule(r){2-3}\cmidrule(l){4-5}
        Model & Agg. & Kmeans & Agg. & Kmeans \\
        \midrule
        AARI & 0.48\std{0.47} & 0.06\std{0.79} & -0.35\std{0.60} & 0.05\std{0.37} \\
        ANMI & 0.57\std{0.46} & 0.04\std{0.78} & -0.31\std{0.54} & 0.10\std{0.46} \\
        CHI & 0.36\std{0.51} & 0.50\std{0.54} &\textbf{0.95\std{0.11}} & -0.02\std{0.40} \\
        SI & 0.07\std{0.69} & \textbf{0.68\std{0.40}} &-0.45\std{0.36} & 0.25\std{0.56} \\
        \cmidrule{2-5}
        KL & 0.38\std{0.63} & 0.05\std{0.79} &-0.52\std{0.44} & 0.05\std{0.37} \\
        Binary & \textbf{0.73\std{0.24}} & 0.02\std{0.91} &0.44\std{0.65} & 0.06\std{0.46} \\
        \bottomrule
    \end{tabular}
\end{table}

%% file: figures/table_benchmark_fixed_regretari_restricted.tex
\begin{table}[t]
    \centering
    \caption{Regret on the ARI of the model selected\std{std} ($\downarrow$) by each ranking metric when the base clusterings are restricted to as many clusters as targets.}
    \label{tab:benchmark_fixed_regretari_restricted}
    \begin{tabular}{lllll}
        \toprule
        &\multicolumn{2}{c}{FCPS}&\multicolumn{2}{c}{UCI}\\
        \cmidrule(r){2-3}\cmidrule(l){4-5}
        Model & Agg. & Kmeans & Agg. & Kmeans \\
        \midrule
        AARI & 0.16\std{0.28} & 0.06\std{0.10} &0.31\std{0.22} & 0.05\std{0.05} \\
        ANMI & 0.16\std{0.28} & 0.06\std{0.10} &0.31\std{0.22} & 0.03\std{0.03} \\
        CHI & 0.24\std{0.33} & 0.05\std{0.09} &\textbf{0.00\std{0.00}} & 0.06\std{0.05} \\
        SI & 0.31\std{0.37} & 0.02\std{0.04} &0.38\std{0.25} & 0.07\std{0.08} \\
        \cmidrule{2-5}
        KL & 0.26\std{0.37} & 0.06\std{0.10} &0.32\std{0.21} & 0.05\std{0.05} \\
        Binary & 0.15\std{0.24} & 0.05\std{0.10} &0.17\std{0.18} & 0.05\std{0.06} \\
        \bottomrule
    \end{tabular}
\end{table}

%% file: figures/table_benchmark_mixed_kendall_restricted.tex
\begin{table*}[t]
    \centering
    \caption{Average Kendall tau correlation\std{std} ($\uparrow$) of the ranking metrics when the base clusterings seek different number of clusters.}
    \label{tab:benchmark_mixed_kendall_restricted}
    \begin{tabular}{lllllllll}
        \toprule
        &\multicolumn{4}{c}{FCPS}&\multicolumn{4}{c}{UCI}\\
        \cmidrule(r){2-5}\cmidrule(l){6-9}
        Model & Agg. & DBSCAN & Kmeans & Alltogether& Agg. & DBSCAN & Kmeans & Alltogether \\
        \midrule
        AARI & 0.34\std{0.32} & 0.62\std{0.48} & 0.20\std{0.36} & 0.28\std{0.42} &0.18\std{0.33} & 0.19\std{0.55} & 0.17\std{0.30} & 0.32\std{0.30} \\
        ANMI & 0.17\std{0.39} & 0.63\std{0.49} & 0.03\std{0.46} & 0.14\std{0.47} &0.30\std{0.37} & 0.16\std{0.53} & -0.01\std{0.38} & 0.36\std{0.22} \\
        CHI & 0.01\std{0.36} & 0.35\std{0.77} & 0.15\std{0.51} & -0.04\std{0.52} &0.39\std{0.24} & 0.40\std{0.27} & 0.33\std{0.46} & 0.36\std{0.26} \\
        WGSS & -0.41\std{0.32} & -0.11\std{0.93} & -0.68\std{0.26} & -0.47\std{0.32} &0.29\std{0.54} & \textbf{0.55\std{0.36}} & -0.37\std{0.49} & 0.20\std{0.38} \\
        \cmidrule{2-9}
        KL & 0.25\std{0.45} & 0.57\std{0.57} & -0.16\std{0.46} & 0.20\std{0.50} &0.06\std{0.40} & 0.17\std{0.51} & -0.11\std{0.39} & 0.31\std{0.38} \\
        Binary & \textbf{0.79\std{0.24}} & \textbf{0.84\std{0.24}} & \textbf{0.82\std{0.09}} & \textbf{0.73\std{0.36}} &\textbf{0.63\std{0.20}} & \textbf{0.55\std{0.30}} & \textbf{0.47\std{0.41}} & \textbf{0.52\std{0.21}} \\
        \bottomrule
    \end{tabular}
\end{table*}

%% file: figures/table_benchmark_mixed_regretari_restricted.tex
\begin{table*}[t]
    \centering
    \caption{Regret on the ARI of the model selected\std{std} ($\downarrow$) by each ranking metric when the base clusterings seek different number of clusters.}
    \label{tab:benchmark_mixed_regretari_restricted}
    \begin{tabular}{lllllllll}
        \toprule
        &\multicolumn{4}{c}{FCPS}&\multicolumn{4}{c}{UCI}\\
        \cmidrule(r){2-5}\cmidrule(l){6-9}
        Model & Agg. & DBSCAN & Kmeans & Alltogether& Agg. & DBSCAN & Kmeans & Alltogether \\
        \midrule
            AARI & 0.41\std{0.32} & 0.10\std{0.17} & 0.33\std{0.27} & 0.41\std{0.32} &0.20\std{0.24} & 0.11\std{0.13} & 0.20\std{0.19} & 0.20\std{0.25} \\
            ANMI & 0.41\std{0.33} & 0.10\std{0.17} & 0.33\std{0.30} & 0.40\std{0.34} &0.17\std{0.23} & 0.12\std{0.13} & 0.19\std{0.24} & 0.14\std{0.15} \\
            CHI & 0.24\std{0.32} & 0.07\std{0.15} & 0.16\std{0.19} & 0.27\std{0.34} &0.13\std{0.20} & 0.06\std{0.09} & 0.15\std{0.19} & 0.12\std{0.17} \\
            SI & 0.24\std{0.34} & 0.08\std{0.17} & \textbf{0.10\std{0.14}} & 0.31\std{0.34} &0.40\std{0.26} & 0.19\std{0.19} & 0.14\std{0.24} & 0.41\std{0.28} \\
        \cmidrule{2-9}
            KL & 0.40\std{0.32} & 0.10\std{0.17} & 0.42\std{0.28} & 0.43\std{0.31} &0.23\std{0.25} & 0.12\std{0.13} & 0.23\std{0.23} & 0.15\std{0.21} \\
            Binary & \textbf{0.14\std{0.20}} & \textbf{0.06\std{0.14}} & 0.17\std{0.21} & \textbf{0.22\std{0.24}} &\textbf{0.11\std{0.16}} & \textbf{0.05\std{0.06}} & \textbf{0.09\std{0.12}} & \textbf{0.10\std{0.14}} \\
        \bottomrule
    \end{tabular}
\end{table*}

%% file: figures/fig_regularisation.tex
\begin{figure}[t]
	\centering
    \includegraphics[width=0.8\linewidth]{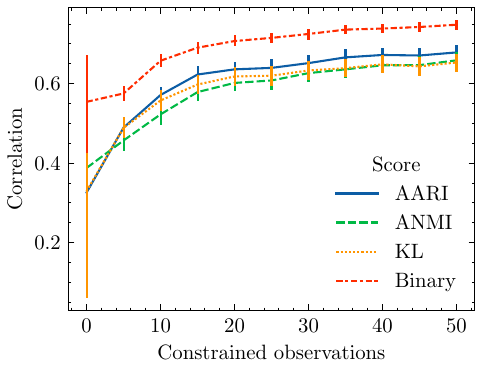}
	\caption{Kendall's tau correlation ($\uparrow$) after adding must-link cannot-link constraints from an increasing number of random observations for the UCI datasets with mixed clustering models.}
	\label{fig:benchmark_correlation_gain}
\end{figure}

%% file: sections/5_conclusion.tex
We introduced a metric based on the distance between connectivity and consensus matrices to rank clustering algorithms, called the DISCOTEC. Overall, this metric works as intended and tends to select clustering models that are most similar to the consensus. We therefore suggest, as validated through experiments, that a diverse pool of clustering algorithms is required to get the most out of the DISCOTEC.

We have shown experimentally that among several choices of distances, the most efficient is to binarise the consensus matrix with respect to its mean and compute its difference with the connectivity matrix. In general, the resulting performance is equal to or better than other ensemble clustering baselines such as the average ARI. The main difference with this baseline is that the DISCOTEC is faster to compute with respect to the number of models. Compared to other internal metrics, the advantage of the DISCOTEC is its tolerance to any type of clustering algorithm, \ie definition of clusters. Consequently, the DISCOTEC shows better performance when the ranking a diverse set of clustering algorithms. In the case of a single clustering algorithm with limited parameters, a specialised internal metric may be preferred.

Finally, we have shown that the DISCOTEC can be regularised with must-link/cannot-link constraints thanks to the approximate measure of informativeness. Moreover, both methods are compatible from a dimensional analysis perspective because they average differences between edges of connectivity matrices.

In future work, it would be interesting to investigate how to further improve the performance of the DISCOTEC when the pool of base clusterings is not diverse. Additionally, it would be interesting to explore different approaches to the raw binarisation of the consensus matrix, \eg a nonlinear bijection to obtain extreme values without being binary.

%% file: appendix/benchmark_details.tex
We used for our experiments two different types of datasets: synthetic datasets from the fundamental clustering problem suite (FCPS) and real datasets from the UCI repository. Their characteristics are summarised in Table~\ref{tab:benchmark_datasets}.

\input{figures/table_datasets}
\input{figures/table_lexicon}

We used different metrics for evaluating the clustering models as baselines. We essentially took metrics available in the permetrics Python package~\citep{thieu_permetrics_2024}, and expanded with additional metrics described in some R packages~\citep{desgraupes_clustering_2013, charrad_nbclust_2014}, which we implemented ourselves. The name of the metrics thus mainly follow the permetrics name, and we propose a lexicon in Table~\ref{tab:benchmark_lexicon}.

%% file: figures/table_datasets.tex
\begin{table}[th]
	\centering
	\caption{Datasets used in experiments}
	\label{tab:benchmark_datasets}
    \hfill
	\subfloat[FCPS datasets]{\begin{tabular}{lrrr}
		\toprule
		Name & $n$ & $d$ & $K$ (Clusters) \\
		\midrule
		Atom & 800 & 3 & 2\\
		Chainlink & 1000 & 3 & 2 \\
		EngyTime & 4096 & 2 & 2 \\
		Hepta & 212 & 3 & 7 \\
		LSun3D & 404 & 3 & 4 \\
		Target & 770 & 2 & 6 \\
		Tetra & 400 & 3 & 4 \\
		TwoDiamonds & 800 & 2 & 2 \\
		WingNut & 1016 & 2 & 2 \\
        \bottomrule
    \end{tabular}}\hfill
	\subfloat[UCI datasets]{\begin{tabular}{lrrr}
		\toprule    
		Name & $n$ & $d$ & $K$ (Classes) \\
        \midrule
        Dermatology&366&33&6\\
        Digits & 1797 & 64 & 10 \\
        Glass & 214 & 9 & 6 \\
        Ionosphere & 351 & 34 & 2 \\
        Iris & 150 & 4 & 3 \\
        Lung & 32 & 54 & 3 \\
        Segmentation & 2310 & 19 & 7 \\
        WDBC & 569 & 30 & 2 \\
        Wine & 178 & 13 & 3 \\
		\bottomrule
	\end{tabular}}
    \hfill
\end{table}

%% file: figures/table_lexicon.tex
\begin{table}[th]
    \centering
    \caption{Lexicon for the name of the metrics used in experiments. The accronyms are taken from the permetrics library~\citep{thieu_permetrics_2024}. Scores marked with an asterisk were re-implemented.}
    \label{tab:benchmark_lexicon}
    \begin{tabular}{lll}
        \toprule
        Short Name & Name & Reference \\
        \midrule
        AARI& Average ARI & -\\
        ANMI & Average NMI & \citealp{strehl_cluster_2002}\\
        BHI & Ball-Hall index & \citealp{ball_isodata_1965}\\
        BRI & Banfield-Raftery index & \citealp{banfield_model-based_1993}\\
        CHI & Calinski-Harabasz index&\citealp{calinski_dendrite_1974}\\
        CI\textsuperscript{*} & C-index & \citealp{hubert_quadratic_1976}\\
        DBI& Davies-Bouldin index&\citealp{davies_cluster_1979} \\
        DHI& Duda-Hart index &\citealp{duda_pattern_1974} \\
        DI\textsuperscript{*}&Dunn index&\citealp{dunn_well-separated_1974} \\
        DRI&Det-ratio index &\citealp{scott_clustering_1971} \\
        HI&Hartigan index&\citealp{hartigan_clustering_1975} \\
        KDI&K squared determinant index&\citealp{marriott_practical_1971} \\
        LDRI&Log det ratio&\citealp{scott_clustering_1971} \\
        LSRI&Log sum of squared error&-\\
        McRao\textsuperscript{*}& McClain-Rao index&\citealp{mcclain_clustisz_1975}\\
        PBM\textsuperscript{*}&Pakhira-Bandyopadhyay-Maulik index&\citealp{pakhira_validity_2004}\\
        SI&Silhouette index&\citealp{rousseeuw_silhouettes_1987}\\
        WGSS&Within-group sum of squares&\citealp{edwards_method_1965}\\
        XBI&Xie-Beni index& \citealp{xie_validity_1991}\\
        WG\textsuperscript{*}& Wermmert-Gancarski index&\citealp{desgraupes_clustering_2013}\\
        \bottomrule
    \end{tabular}
\end{table}

%% file: appendix/scaling_experiment.tex
In this experiment, we further explore the relationship between the number models and the quality of the ranking. We keep the initial synthetic scenario from our first experiment in Section~\ref{sec:experiment}, where a ground truth is first generated and then $T$ models are created by preserving between 10\% and $\rho_\text{max}$ of the labels. The resulting models have an accuracy bounded between 10\% and $\rho_\text{max}$ on average. For three specific thresholds $\rho_\text{max}\in \{0.2, 0.5, 0.9\}$, which correspond to a decreasing difficulty of consensus, we increase the number of models $T$ from 5 models to 200. We report the average correlations for 50 runs per value of $T$ and $\rho_\text{max}$. Figure~\ref{fig:scaling_pearson} corresponds to the Pearson correlation and Figure~\ref{fig:scaling_kendall} corresponds to Kendall's tau correlation coefficient. For clarity of both figures, we omitted the squared Hellinger and total variation distances because they perfectly followed the KL curve.

\input{figures/fig_scaling_pearson}
\input{figures/fig_scaling_kendall}

We observe from both figures that the scaling depends on the difficulty to reach a consensus. When a consensus is hard to find, \ie $\rho_\text{max}=0.2$, even 200 models is insufficient to establish a strong correlation between ranking and ARI with targets. In contrast, an easy scenario, \ie $\rho_\text{max}=0.9$, requires few models to achieve excellent correlations as we are already close to 1 with 20 models. In a mitigated scenario, increasing the number of models increases steadily the correlations. It is notable that the binary DISCOTEC displays stronger performances even in a mitigated scenario compared to the DISCOTEC based on the KL distance.

We conclude that when the consensus is not clear-cut, adding models in the ensemble may be beneficial to the DISCOTEC ranking.

%% file: figures/fig_scaling_pearson.tex
\begin{figure}[th]
    \centering
    \subfloat[$\rho_\text{max} = 0.2$]{
        \includegraphics[width=0.3\linewidth]{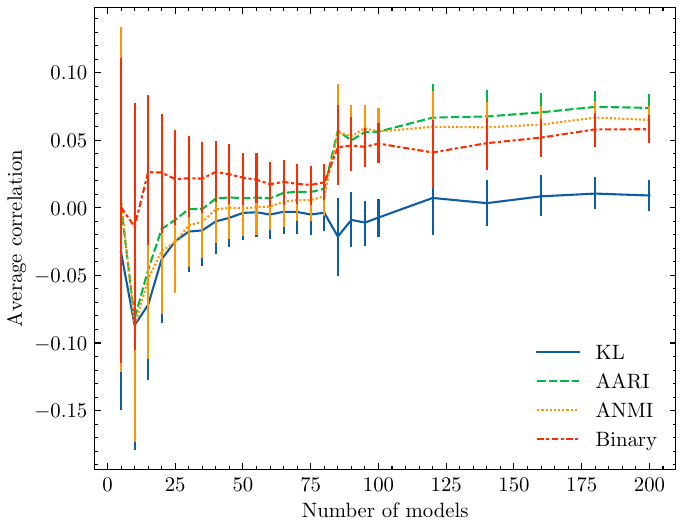}
    }
    \subfloat[$\rho_\text{max} = 0.5$]{
        \includegraphics[width=0.3\linewidth]{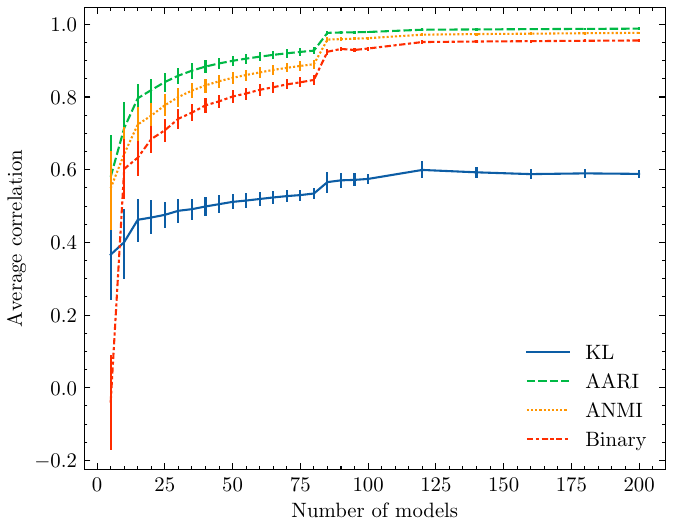}
    }
    \subfloat[$\rho_\text{max} = 0.9$]{
        \includegraphics[width=0.3\linewidth]{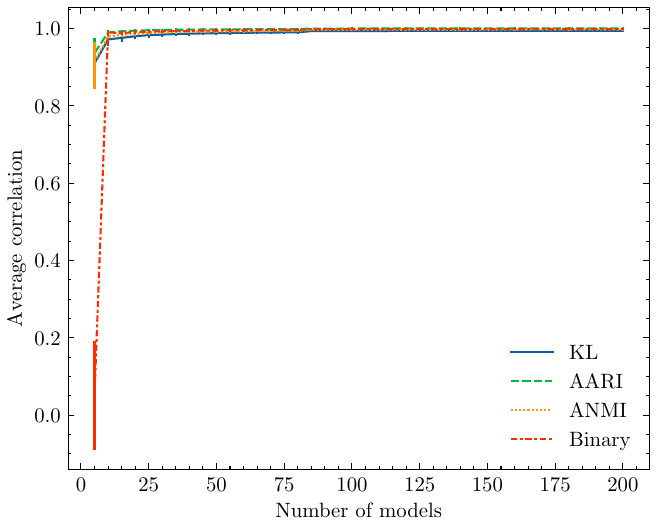}
    }
    \caption{Pearson correlation between ranking metrics and ARI with ground truth as the number of models in the ensemble increase. Each model has an accuracy bounded between 10\% and $\rho_\text{max}$.}
    \label{fig:scaling_pearson}
\end{figure}

%% file: figures/fig_scaling_kendall.tex
\begin{figure}[th]
    \centering
    \subfloat[$\rho_\text{max} = 0.2$]{
        \includegraphics[width=0.3\linewidth]{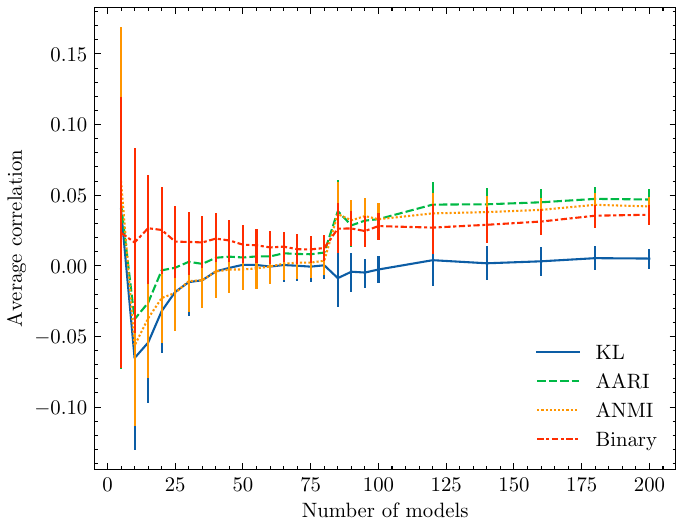}
    }
    \subfloat[$\rho_\text{max} = 0.5$]{
        \includegraphics[width=0.3\linewidth]{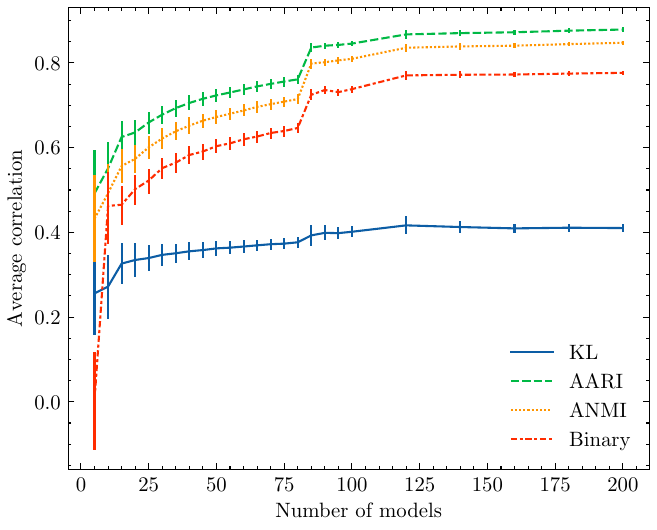}
    }
    \subfloat[$\rho_\text{max} = 0.9$]{
        \includegraphics[width=0.3\linewidth]{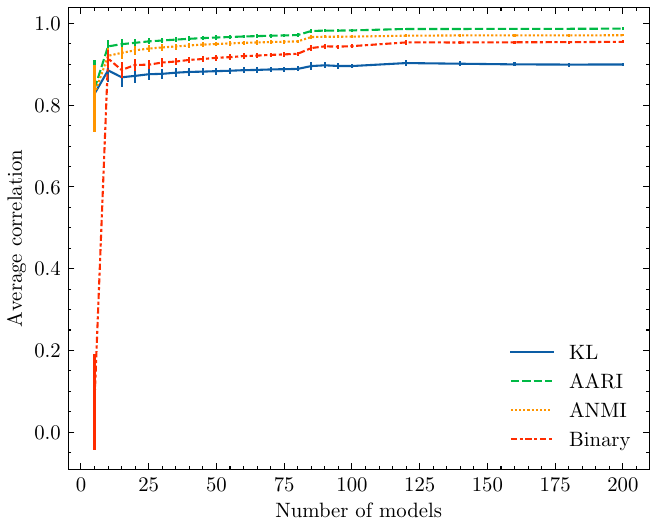}
    }
    \caption{Kendall's tau correlation between ranking metrics and ARI with ground truth as the number of models in the ensemble increase. Each model has an accuracy bounded between 10\% and $\rho_\text{max}$.}
    \label{fig:scaling_kendall}
\end{figure}

%% file: appendix/consensus_visualisation.tex
We show in Figure~\ref{fig:synthetic_consensus} the examples of 3 consensus matrices from the first synthetic scenario in Section~\ref{sec:experiment}.

\input{figures/fig_synthetic_partitions_consensus}

%% file: figures/fig_synthetic_partitions_consensus.tex
\begin{figure}[th]
    \centering
    \includegraphics[width=\linewidth]{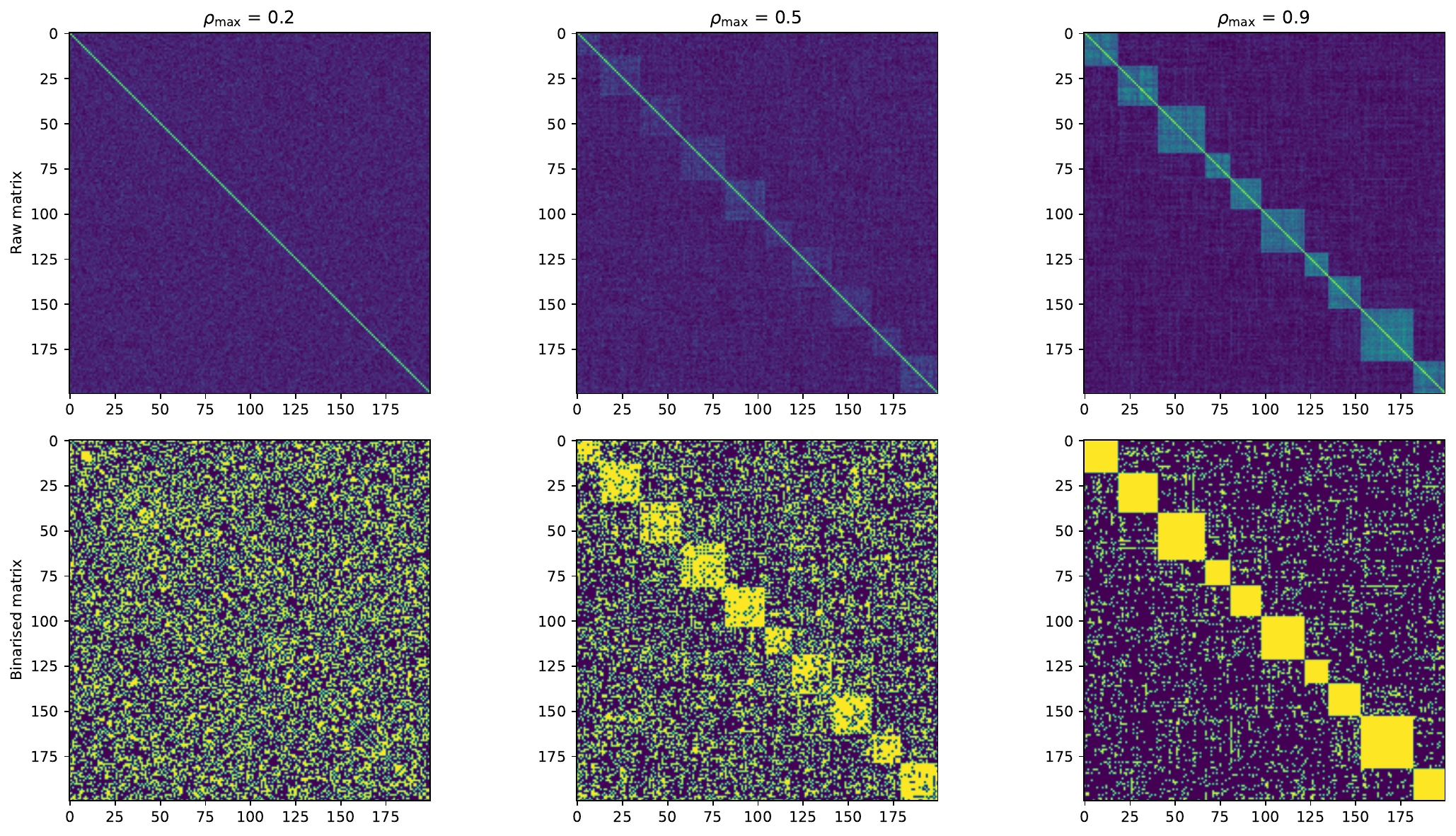}
    \caption{Example of consensus matrices from the first synthetic scenario when the upper bound on the label preservation rate $\rho_\text{max}$ increases from 20\% to 90\% with the ground truth. The top row is shows the initial consensus matrix, the bottom row shows the same matrix after binarising with respect to its mean value.}
    \label{fig:synthetic_consensus}
\end{figure}

%% file: appendix/extended_benchmark.tex
This section describes all extended tables of the experiments from Section~\ref{sec:experiment}. They concern the DISCOTEC performances and all baselines when testing different clustering algorithms on both the FCPS and UCI dataset.

\subsection{Complete tables with Kendall's tau correlation}
\label{appsec:extended_benchmark_kendall}

The tables~\ref{tab:benchmark_fixed_kendall_extended} and \ref{tab:benchmark_fixed_regretkendall_extended} correspond to the performances of scoring methods when the base clusterings are restricted to as many clusters as the number specified by the targets of the dataset.

\input{figures/table_benchmark_fixed_kendall_extended}
\input{figures/table_benchmark_fixed_regretkendall_extended}

The tables~\ref{tab:benchmark_mixed_kendall_extended} and~\ref{tab:benchmark_mixed_regretkendall_extended} correspond to the extension of the experiment where clustering models cover different number of clusters.

\input{figures/table_benchmark_mixed_kendall_extended}

\input{figures/table_benchmark_mixed_regretkendall_extended}

\subsection{Additional tables with Pearson correlation}
\label{appsec:extended_benchmark_pearson}

To further complete the results, we show the exact same tables measuring the Pearson correlation instead of the Kendall's tau correlation. The tables~\ref{tab:benchmark_fixed_pearson_extended} and~\ref{tab:benchmark_fixed_regretpearson_extended} respectively correspond to the correlation and regret on correlation when the number of clusters is fixed. The tables~\ref{tab:benchmark_mixed_pearson_extended} and~\ref{tab:benchmark_mixed_regretpearson_extended} respectively correspond to the correlation and regret on correlation when the number of clusters varies between clustering algorithms.

\input{figures/table_benchmark_fixed_pearson_extended}
\input{figures/table_benchmark_fixed_regretpearson_extended}
\input{figures/table_benchmark_mixed_pearson_extended}

\input{figures/table_benchmark_mixed_regretpearson_extended}

\subsection{Complete tables for the ARI regret of selected models}
\label{appsec:extended_benchmark_ari}

We finally report the extended results for the regret on the ARI of the top-selected model per score in Table~\ref{tab:benchmark_fixed_regretari_extended} when the number of clusters is restricted and Talbe~\ref{tab:benchmark_mixed_regretari_extended} when the number of clusters can vary.

\input{figures/table_benchmark_mixed_regretari_extended}
\input{figures/table_benchmark_fixed_regretari_extended}

%% file: figures/table_benchmark_fixed_kendall_extended.tex
\begin{table}[th]
    \centering
    \caption{Extended results for the average Kendall tau correlation\std{std} ($\uparrow$) of the ranking metrics when the base clusterings are restricted to as many clusters as targets.}
    \label{tab:benchmark_fixed_kendall_extended}
    \resizebox{!}{0.8\height}{
    \begin{tabular}{lllll}
        \toprule
        &\multicolumn{2}{c}{FCPS}&\multicolumn{2}{c}{UCI}\\
        \cmidrule(r){2-3}\cmidrule(l){4-5}
        Model & Agglomerative & Kmeans & Agglomerative & Kmeans \\
        \midrule
        AARI & 0.48\std{0.47} & 0.06\std{0.79} & -0.35\std{0.60} & 0.05\std{0.37} \\
        ANMI & 0.57\std{0.46} & 0.04\std{0.78} & -0.31\std{0.54} & 0.10\std{0.46} \\
        BHI & 0.36\std{0.51} & 0.47\std{0.54} &\textbf{0.95\std{0.11}} & -0.02\std{0.40} \\
        BRI & 0.63\std{0.37} & 0.41\std{0.58} &0.86\std{0.13} & 0.21\std{0.60} \\
        CHI & 0.36\std{0.51} & 0.50\std{0.54} &\textbf{0.95\std{0.11}} & -0.02\std{0.40} \\
        CI & 0.39\std{0.37} & 0.41\std{0.52} &-0.35\std{0.46} & -0.21\std{0.45} \\
        DBI & 0.02\std{0.77} & 0.25\std{0.69} &-0.80\std{0.25} & 0.27\std{0.38} \\
        DHI & -0.08\std{0.54} & 0.45\std{0.56} &-0.80\std{0.19} & 0.16\std{0.59} \\
        DI & 0.36\std{0.58} & 0.10\std{0.73} &-0.68\std{0.36} & 0.08\std{0.24} \\
        DRI & 0.21\std{0.73} & 0.42\std{0.64} &0.55\std{0.66} & 0.29\std{0.34} \\
        HI & 0.02\std{0.79} & 0.42\std{0.55} &-0.83\std{0.20} & 0.33\std{0.42} \\
        KDI & -0.16\std{0.56} & 0.19\std{0.62} &0.02\std{0.56} & -0.01\std{0.53} \\
        LDRI & 0.21\std{0.73} & 0.42\std{0.64} &0.55\std{0.66} & 0.29\std{0.34} \\
        LSRI & 0.36\std{0.51} & 0.52\std{0.55} &\textbf{0.95\std{0.11}} & -0.02\std{0.40} \\
        McRao & 0.26\std{0.43} & 0.38\std{0.47} &-0.30\std{0.56} & -0.18\std{0.37} \\
        PBM & -0.02\std{0.63} & 0.41\std{0.69} &-0.32\std{0.59} & 0.09\std{0.63} \\
        SI & 0.07\std{0.69} & \textbf{0.68\std{0.40}} &-0.45\std{0.36} & 0.25\std{0.56} \\
        WGSS & 0.36\std{0.51} & 0.51\std{0.58} &\textbf{0.95\std{0.11}} & -0.02\std{0.40} \\
        WG & -0.25\std{0.62} & 0.39\std{0.57} &-0.93\std{0.12} & 0.04\std{0.47} \\
        XBI & 0.12\std{0.67} & 0.42\std{0.60} &-0.61\std{0.20} & 0.14\std{0.52} \\
        \cmidrule{2-5}
        H & 0.42\std{0.64} & 0.05\std{0.79} &-0.52\std{0.44} & 0.05\std{0.37} \\
        KL & 0.38\std{0.63} & 0.05\std{0.79} &-0.52\std{0.44} & 0.05\std{0.37} \\
        TV & 0.42\std{0.64} & 0.05\std{0.79} &-0.52\std{0.44} & 0.05\std{0.37} \\
        Binary & \textbf{0.73\std{0.24}} & 0.02\std{0.91} &0.44\std{0.65} & 0.06\std{0.46} \\
        \bottomrule
    \end{tabular}}
\end{table}

%% file: figures/table_benchmark_fixed_regretkendall_extended.tex
\begin{table}[th]
    \centering
    \caption{Extended results for the regret on the Kendall tau correlation\std{std} ($\downarrow$) of the ranking metrics when the base clusterings are restricted to as many clusters as targets.}
    \label{tab:benchmark_fixed_regretkendall_extended}
    \resizebox{!}{0.8\height}{
    \begin{tabular}{lllll}
        \toprule
        &\multicolumn{2}{c}{FCPS}&\multicolumn{2}{c}{UCI}\\
        \cmidrule(r){2-3}\cmidrule(l){4-5}
        Model & Agglomerative & Kmeans & Agglomerative & Kmeans \\
        \midrule
        AARI & 0.40\std{0.48} & 0.78\std{0.84} &1.31\std{0.65} & 0.62\std{0.50} \\
        ANMI & 0.30\std{0.48} & 0.81\std{0.82} &1.28\std{0.59} & 0.57\std{0.54} \\
        BHI & 0.52\std{0.49} & 0.37\std{0.47} &\textbf{0.01\std{0.03}} & 0.69\std{0.53} \\
        BRI & 0.24\std{0.28} & 0.43\std{0.54} &0.11\std{0.13} & 0.46\std{0.60} \\
        CHI & 0.52\std{0.49} & 0.34\std{0.46} &\textbf{0.01\std{0.03}} & 0.69\std{0.53} \\
        CI & 0.49\std{0.33} & 0.43\std{0.45} &1.31\std{0.52} & 0.88\std{0.66} \\
        DBI & 0.86\std{0.81} & 0.60\std{0.69} &1.77\std{0.33} & 0.40\std{0.22} \\
        DHI & 0.95\std{0.59} & 0.39\std{0.55} &1.77\std{0.24} & 0.51\std{0.45} \\
        DI & 0.52\std{0.58} & 0.74\std{0.75} &1.65\std{0.43} & 0.59\std{0.38} \\
        DRI & 0.67\std{0.70} & 0.42\std{0.56} &0.39\std{0.69} & 0.47\std{0.14} \\
        HI & 0.85\std{0.84} & 0.43\std{0.53} &1.80\std{0.25} & 0.34\std{0.19} \\
        KDI & 1.03\std{0.54} & 0.65\std{0.63} &0.94\std{0.52} & 0.70\std{0.66} \\
        LDRI & 0.67\std{0.70} & 0.42\std{0.56} &0.39\std{0.69} & 0.47\std{0.14} \\
        LSRI & 0.52\std{0.49} & 0.33\std{0.46} &\textbf{0.01\std{0.03}} & 0.69\std{0.53} \\
        McRao & 0.61\std{0.46} & 0.46\std{0.42} &1.27\std{0.58} & 0.85\std{0.59} \\
        PBM & 0.89\std{0.68} & 0.44\std{0.62} &1.29\std{0.62} & 0.59\std{0.64} \\
        SI & 0.81\std{0.73} & \textbf{0.16\std{0.23}} &1.42\std{0.45} & 0.42\std{0.47} \\
        WGSS & 0.52\std{0.49} & 0.34\std{0.50} &\textbf{0.01\std{0.03}} & 0.69\std{0.53} \\
        WG & 1.14\std{0.64} & 0.45\std{0.57} &1.90\std{0.21} & 0.63\std{0.43} \\
        XBI & 0.76\std{0.72} & 0.42\std{0.51} &1.58\std{0.26} & 0.54\std{0.59} \\
        \cmidrule{2-5}
        H & 0.46\std{0.70} & 0.79\std{0.83} &1.49\std{0.52} & 0.62\std{0.50} \\
        KL & 0.50\std{0.68} & 0.79\std{0.83} &1.49\std{0.52} & 0.62\std{0.50} \\
        TV & 0.46\std{0.70} & 0.79\std{0.83} &1.49\std{0.52} & 0.62\std{0.50} \\
        Binary & \textbf{0.15\std{0.21}} & 0.82\std{0.96} &0.52\std{0.65} & 0.62\std{0.50} \\
        \bottomrule
    \end{tabular}}
\end{table}

%% file: figures/table_benchmark_mixed_kendall_extended.tex
\begin{table}[th]
    \centering
    \caption{Extended results for the average Kendall tau correlation\std{std} ($\uparrow$) of the ranking metrics when the base clusterings seek different number of clusters.}
    \label{tab:benchmark_mixed_kendall_extended}
    \resizebox{!}{0.8\height}{
    \begin{tabular}{lllllllll}
        \toprule
        &\multicolumn{4}{c}{FCPS}&\multicolumn{4}{c}{UCI}\\
        \cmidrule(r){2-5}\cmidrule(l){6-9}
        Model & Agg. & DBSCAN & Kmeans & Alltogether& Agg. & DBSCAN & Kmeans & Alltogether \\
        \midrule
        AARI & 0.34\std{0.32} & 0.62\std{0.48} & 0.20\std{0.36} & 0.28\std{0.42} &0.18\std{0.33} & 0.19\std{0.55} & 0.17\std{0.30} & 0.32\std{0.30} \\
        ANMI & 0.17\std{0.39} & 0.63\std{0.49} & 0.03\std{0.46} & 0.14\std{0.47} &0.30\std{0.37} & 0.16\std{0.53} & -0.01\std{0.38} & 0.36\std{0.22} \\
        BHI & -0.34\std{0.29} & -0.20\std{0.83} & -0.68\std{0.26} & -0.43\std{0.30} &0.12\std{0.45} & 0.47\std{0.39} & -0.37\std{0.49} & 0.10\std{0.37} \\
        BRI & -0.40\std{0.31} & -0.19\std{0.81} & -0.69\std{0.25} & -0.43\std{0.31} &0.19\std{0.20} & 0.33\std{0.61} & -0.07\std{0.26} & 0.23\std{0.23} \\
        CHI & 0.01\std{0.36} & 0.35\std{0.77} & 0.15\std{0.51} & -0.04\std{0.52} &0.39\std{0.24} & 0.40\std{0.27} & 0.33\std{0.46} & 0.36\std{0.26} \\
        CI & -0.06\std{0.42} & -0.15\std{0.89} & -0.35\std{0.44} & -0.21\std{0.48} &-0.12\std{0.40} & 0.07\std{0.57} & -0.12\std{0.43} & 0.04\std{0.29} \\
        DBI & 0.03\std{0.38} & -0.23\std{0.78} & 0.02\std{0.37} & -0.10\std{0.40} &-0.38\std{0.33} & -0.19\std{0.34} & 0.03\std{0.29} & -0.08\std{0.24} \\
        DHI & -0.09\std{0.26} & -0.35\std{0.62} & -0.66\std{0.26} & -0.29\std{0.26} &-0.37\std{0.15} & 0.01\std{0.43} & -0.31\std{0.43} & -0.17\std{0.23} \\
        DI & 0.11\std{0.27} & 0.30\std{0.71} & -0.11\std{0.21} & 0.09\std{0.27} &-0.48\std{0.27} & -0.17\std{0.57} & -0.01\std{0.29} & -0.18\std{0.31} \\
        DRI & -0.42\std{0.33} & -0.12\std{0.92} & -0.68\std{0.26} & -0.48\std{0.33} &0.04\std{0.50} & 0.47\std{0.30} & -0.49\std{0.39} & 0.08\std{0.29} \\
        HI & 0.28\std{0.36} & 0.26\std{0.65} & 0.62\std{0.28} & 0.29\std{0.34} &-0.29\std{0.49} & -0.36\std{0.48} & 0.34\std{0.53} & -0.19\std{0.42} \\
        KDI & 0.16\std{0.25} & 0.29\std{0.68} & 0.63\std{0.26} & 0.30\std{0.21} &-0.11\std{0.37} & 0.07\std{0.51} & 0.35\std{0.31} & -0.03\std{0.37} \\
        LDRI & -0.42\std{0.33} & -0.12\std{0.92} & -0.68\std{0.26} & -0.48\std{0.33} &0.04\std{0.50} & 0.44\std{0.38} & -0.49\std{0.39} & 0.05\std{0.31} \\
        LSRI & -0.41\std{0.32} & -0.11\std{0.93} & -0.68\std{0.26} & -0.47\std{0.32} &0.29\std{0.54} & \textbf{0.55\std{0.36}} & -0.37\std{0.49} & 0.20\std{0.38} \\
        McRao & -0.39\std{0.27} & -0.28\std{0.81} & -0.66\std{0.28} & -0.47\std{0.30} &-0.28\std{0.45} & -0.02\std{0.63} & -0.30\std{0.46} & -0.12\std{0.38} \\
        PBM & 0.12\std{0.35} & 0.21\std{0.74} & 0.47\std{0.47} & 0.14\std{0.47} &-0.01\std{0.29} & -0.22\std{0.63} & 0.38\std{0.26} & 0.01\std{0.29} \\
        SI & 0.07\std{0.32} & 0.17\std{0.72} & 0.32\std{0.36} & 0.01\std{0.44} &-0.30\std{0.37} & -0.18\std{0.62} & 0.23\std{0.32} & -0.11\std{0.26} \\
        WGSS & -0.41\std{0.32} & -0.11\std{0.93} & -0.68\std{0.26} & -0.47\std{0.32} &0.29\std{0.54} & \textbf{0.55\std{0.36}} & -0.37\std{0.49} & 0.20\std{0.38} \\
        WG & 0.04\std{0.24} & -0.38\std{0.62} & -0.45\std{0.32} & -0.18\std{0.28} &-0.19\std{0.18} & 0.10\std{0.46} & -0.21\std{0.41} & -0.16\std{0.22} \\
        XBI & -0.12\std{0.33} & 0.14\std{0.78} & 0.09\std{0.27} & -0.12\std{0.34} &-0.29\std{0.34} & -0.23\std{0.60} & 0.14\std{0.26} & -0.16\std{0.27} \\
        \cmidrule{2-9}
        H & 0.24\std{0.46} & 0.57\std{0.57} & -0.16\std{0.46} & 0.20\std{0.50} &0.06\std{0.41} & 0.18\std{0.51} & -0.11\std{0.39} & 0.31\std{0.38} \\
        KL & 0.25\std{0.45} & 0.57\std{0.57} & -0.16\std{0.46} & 0.20\std{0.50} &0.06\std{0.40} & 0.17\std{0.51} & -0.11\std{0.39} & 0.31\std{0.38} \\
        TV & 0.24\std{0.46} & 0.57\std{0.57} & -0.16\std{0.46} & 0.19\std{0.50} &0.06\std{0.41} & 0.18\std{0.51} & -0.11\std{0.39} & 0.31\std{0.39} \\
        Binary & \textbf{0.79\std{0.24}} & \textbf{0.84\std{0.24}} & \textbf{0.82\std{0.09}} & \textbf{0.73\std{0.36}} &\textbf{0.63\std{0.20}} & \textbf{0.55\std{0.30}} & \textbf{0.47\std{0.41}} & \textbf{0.52\std{0.21}} \\
        \bottomrule
    \end{tabular}}
\end{table}

%% file: figures/table_benchmark_mixed_regretkendall_extended.tex
\begin{table}[th]
    \centering
    \caption{Extended results for the regret on the Kendall tau correlation\std{std} ($\downarrow$) of the ranking metrics when the base clusterings seek different number of clusters.}
    \label{tab:benchmark_mixed_regretkendall_extended}
    \resizebox{!}{0.8\height}{
    \begin{tabular}{lllllllll}
        \toprule
        &\multicolumn{4}{c}{FCPS}&\multicolumn{4}{c}{UCI}\\
        \cmidrule(r){2-5}\cmidrule(l){6-9}
        Model & Agg. & DBSCAN & Kmeans & Alltogether& Agg. & DBSCAN & Kmeans & Alltogether \\
        \midrule
        AARI & 0.47\std{0.32} & 0.26\std{0.34} & 0.65\std{0.35} & 0.55\std{0.39} &0.55\std{0.40} & 0.64\std{0.55} & 0.46\std{0.36} & 0.31\std{0.30} \\
        ANMI & 0.64\std{0.42} & 0.25\std{0.36} & 0.83\std{0.46} & 0.69\std{0.48} &0.43\std{0.40} & 0.67\std{0.54} & 0.65\std{0.45} & 0.26\std{0.13} \\
        BHI & 1.15\std{0.41} & 1.10\std{0.86} & 1.54\std{0.30} & 1.25\std{0.39} &0.61\std{0.43} & 0.37\std{0.40} & 1.01\std{0.65} & 0.53\std{0.33} \\
        BRI & 1.21\std{0.43} & 1.09\std{0.86} & 1.54\std{0.29} & 1.25\std{0.42} &0.54\std{0.23} & 0.50\std{0.62} & 0.71\std{0.48} & 0.39\std{0.24} \\
        CHI & 0.80\std{0.35} & 0.55\std{0.74} & 0.71\std{0.47} & 0.86\std{0.51} &0.35\std{0.25} & 0.44\std{0.34} & 0.31\std{0.42} & 0.26\std{0.26} \\
        CI & 0.87\std{0.44} & 1.05\std{0.91} & 1.21\std{0.46} & 1.03\std{0.49} &0.85\std{0.45} & 0.76\std{0.61} & 0.76\std{0.55} & 0.58\std{0.30} \\
        DBI & 0.78\std{0.33} & 1.13\std{0.75} & 0.83\std{0.39} & 0.93\std{0.41} &1.12\std{0.37} & 1.03\std{0.40} & 0.61\std{0.31} & 0.70\std{0.19} \\
        DHI & 0.90\std{0.40} & 1.25\std{0.65} & 1.51\std{0.31} & 1.11\std{0.47} &1.11\std{0.24} & 0.82\std{0.48} & 0.94\std{0.58} & 0.79\std{0.21} \\
        DI & 0.70\std{0.37} & 0.59\std{0.62} & 0.96\std{0.25} & 0.73\std{0.27} &1.22\std{0.31} & 1.00\std{0.60} & 0.65\std{0.43} & 0.81\std{0.40} \\
        DRI & 1.23\std{0.43} & 1.01\std{0.95} & 1.54\std{0.31} & 1.30\std{0.40} &0.69\std{0.56} & 0.33\std{0.28} & 1.12\std{0.72} & 0.57\std{0.31} \\
        HI & 0.53\std{0.28} & 0.63\std{0.59} & 0.24\std{0.26} & 0.53\std{0.38} &1.02\std{0.53} & 1.19\std{0.52} & 0.29\std{0.45} & 0.81\std{0.48} \\
        KDI & 0.65\std{0.17} & 0.61\std{0.64} & 0.23\std{0.24} & 0.53\std{0.26} &0.88\std{0.37} & 0.77\std{0.48} & 0.34\std{0.30} & 0.68\std{0.43} \\
        LDRI & 1.23\std{0.43} & 1.01\std{0.95} & 1.54\std{0.31} & 1.30\std{0.40} &0.69\std{0.56} & 0.36\std{0.36} & 1.12\std{0.72} & 0.60\std{0.29} \\
        LSRI & 1.22\std{0.42} & 1.01\std{0.95} & 1.53\std{0.30} & 1.29\std{0.39} &0.44\std{0.51} & \textbf{0.28\std{0.39}} & 1.01\std{0.65} & 0.43\std{0.35} \\
        McRao & 1.20\std{0.38} & 1.18\std{0.81} & 1.51\std{0.34} & 1.30\std{0.39} &1.01\std{0.49} & 0.85\std{0.68} & 0.94\std{0.62} & 0.74\std{0.39} \\
        PBM & 0.69\std{0.27} & 0.69\std{0.72} & 0.38\std{0.45} & 0.69\std{0.52} &0.74\std{0.34} & 1.05\std{0.68} & 0.26\std{0.16} & 0.61\std{0.33} \\
        SI & 0.74\std{0.31} & 0.72\std{0.63} & 0.54\std{0.35} & 0.81\std{0.44} &1.04\std{0.42} & 1.01\std{0.69} & 0.40\std{0.28} & 0.74\std{0.22} \\
        WGSS & 1.22\std{0.42} & 1.01\std{0.95} & 1.53\std{0.30} & 1.29\std{0.39} &0.44\std{0.51} & \textbf{0.28\std{0.39}} & 1.01\std{0.65} & 0.43\std{0.35} \\
        WG & 0.77\std{0.36} & 1.28\std{0.64} & 1.30\std{0.34} & 1.00\std{0.42} &0.92\std{0.21} & 0.74\std{0.48} & 0.85\std{0.53} & 0.78\std{0.20} \\
        XBI & 0.93\std{0.36} & 0.76\std{0.71} & 0.76\std{0.33} & 0.94\std{0.38} &1.03\std{0.37} & 1.06\std{0.66} & 0.49\std{0.25} & 0.78\std{0.27} \\
        \cmidrule{2-9}
        H & 0.57\std{0.48} & 0.31\std{0.44} & 1.01\std{0.49} & 0.63\std{0.51} &0.67\std{0.49} & 0.65\std{0.52} & 0.74\std{0.52} & 0.31\std{0.39} \\
        KL & 0.56\std{0.48} & 0.31\std{0.44} & 1.01\std{0.48} & 0.62\std{0.51} &0.67\std{0.48} & 0.66\std{0.51} & 0.74\std{0.52} & 0.31\std{0.38} \\
        TV & 0.57\std{0.48} & 0.31\std{0.44} & 1.01\std{0.49} & 0.63\std{0.51} &0.68\std{0.49} & 0.65\std{0.52} & 0.75\std{0.52} & 0.31\std{0.39} \\
        Binary & \textbf{0.02\std{0.07}} & \textbf{0.05\std{0.14}} & \textbf{0.03\std{0.05}} & \textbf{0.09\std{0.25}} &\textbf{0.10\std{0.20}} & 0.29\std{0.31} & \textbf{0.17\std{0.29}} & \textbf{0.11\std{0.13}} \\
        \bottomrule
    \end{tabular}}
\end{table}

%% file: figures/table_benchmark_fixed_pearson_extended.tex
\begin{table}[th]
    \centering
    \caption{Extended results for the average Pearson correlation\std{std} ($\uparrow$) of the ranking metrics when the base clusterings are restricted to as many clusters as targets.}
    \label{tab:benchmark_fixed_pearson_extended}
    \resizebox{!}{0.8\height}{
    \begin{tabular}{lllll}
        \toprule
        &\multicolumn{2}{c}{FCPS}&\multicolumn{2}{c}{UCI}\\
        \cmidrule(r){2-3}\cmidrule(l){4-5}
        Model & Agglomerative & Kmeans & Agglomerative & Kmeans \\
        \midrule
        AARI & 0.72\std{0.49} & 0.03\std{0.94} &-0.35\std{0.78} & 0.43\std{0.54} \\
        ANMI & 0.80\std{0.39} & -0.01\std{0.86} &-0.29\std{0.77} & 0.49\std{0.48} \\
        BHI & 0.41\std{0.73} & 0.57\std{0.68} &0.92\std{0.10} & 0.37\std{0.57} \\
        BRI & 0.63\std{0.28} & 0.33\std{0.83} &0.87\std{0.11} & 0.30\std{0.64} \\
        CHI & 0.41\std{0.69} & 0.56\std{0.68} &\textbf{0.93\std{0.10}} & 0.37\std{0.57} \\
        CI & 0.39\std{0.68} & 0.57\std{0.65} &-0.39\std{0.53} & 0.05\std{0.67} \\
        DBI & 0.15\std{0.79} & 0.48\std{0.69} &-0.78\std{0.17} & 0.02\std{0.63} \\
        DHI & 0.05\std{0.60} & 0.56\std{0.56} &-0.90\std{0.11} & -0.15\std{0.62} \\
        DI & 0.64\std{0.45} & 0.18\std{0.76} &-0.84\std{0.13} & -0.01\std{0.39} \\
        DRI & 0.22\std{0.85} & 0.44\std{0.86} &0.51\std{0.84} & 0.55\std{0.41} \\
        HI & 0.13\std{0.78} & 0.47\std{0.72} &-0.85\std{0.13} & -0.03\std{0.59} \\
        KDI & -0.27\std{0.69} & 0.12\std{0.81} &-0.08\std{0.72} & -0.22\std{0.52} \\
        LDRI & 0.22\std{0.86} & 0.44\std{0.86} &0.55\std{0.86} & \textbf{0.61\std{0.39}} \\
        LSRI & 0.35\std{0.72} & 0.58\std{0.67} &0.92\std{0.09} & 0.37\std{0.57} \\
        McRao & 0.33\std{0.66} & 0.56\std{0.70} &-0.32\std{0.74} & 0.10\std{0.63} \\
        PBM & 0.26\std{0.74} & 0.52\std{0.61} &-0.38\std{0.66} & 0.06\std{0.67} \\
        SI & 0.20\std{0.79} & \textbf{0.67\std{0.56}} &-0.63\std{0.37} & 0.10\std{0.72} \\
        WGSS & 0.41\std{0.73} & 0.57\std{0.68} &0.92\std{0.10} & 0.37\std{0.57} \\
        WG & -0.13\std{0.66} & 0.48\std{0.59} &-0.92\std{0.06} & -0.08\std{0.55} \\
        XBI & 0.23\std{0.80} & 0.59\std{0.56} &-0.66\std{0.25} & 0.22\std{0.60} \\
        \cmidrule{2-5}
        H & 0.53\std{0.71} & 0.04\std{0.92} &-0.51\std{0.68} & 0.47\std{0.51} \\
        KL & 0.53\std{0.70} & 0.04\std{0.91} &-0.51\std{0.68} & 0.47\std{0.48} \\
        TV & 0.53\std{0.72} & 0.03\std{0.92} &-0.51\std{0.68} & 0.46\std{0.52} \\
        Binary & \textbf{0.85\std{0.24}} & 0.01\std{0.97} &0.46\std{0.66} & 0.52\std{0.56} \\
        \bottomrule
    \end{tabular}}
\end{table}

%% file: figures/table_benchmark_fixed_regretpearson_extended.tex
\begin{table}[th]
    \centering
    \caption{Extended results for the regret on the Pearson correlation\std{std} ($\downarrow$) of the ranking metrics when the base clusterings are restricted to as many clusters as targets.}
    \label{tab:benchmark_fixed_regretpearson_extended}
    \resizebox{!}{0.8\height}{
    \begin{tabular}{lllll}
        \toprule
        &\multicolumn{2}{c}{FCPS}&\multicolumn{2}{c}{UCI}\\
        \cmidrule(r){2-3}\cmidrule(l){4-5}
        Model & Agglomerative & Kmeans & Agglomerative & Kmeans \\
        \midrule
        AARI & 0.25\std{0.48} & 0.91\std{0.96} &1.33\std{0.78} & 0.33\std{0.39} \\
        ANMI & 0.17\std{0.39} & 0.95\std{0.87} &1.27\std{0.77} & 0.27\std{0.37} \\
        BHI & 0.55\std{0.71} & 0.37\std{0.67} &\textbf{0.06\std{0.09}} & 0.39\std{0.52} \\
        BRI & 0.34\std{0.27} & 0.61\std{0.84} &0.11\std{0.11} & 0.46\std{0.63} \\
        CHI & 0.55\std{0.68} & 0.38\std{0.67} &\textbf{0.06\std{0.10}} & 0.39\std{0.52} \\
        CI & 0.57\std{0.66} & 0.37\std{0.64} &1.38\std{0.54} & 0.71\std{0.72} \\
        DBI & 0.82\std{0.78} & 0.46\std{0.67} &1.77\std{0.18} & 0.73\std{0.66} \\
        DHI & 0.92\std{0.60} & 0.38\std{0.53} &1.88\std{0.11} & 0.91\std{0.70} \\
        DI & 0.33\std{0.44} & 0.76\std{0.74} &1.83\std{0.14} & 0.76\std{0.48} \\
        DRI & 0.74\std{0.83} & 0.50\std{0.86} &0.48\std{0.85} & 0.33\std{0.29} \\
        HI & 0.84\std{0.77} & 0.47\std{0.70} &1.83\std{0.14} & 0.78\std{0.63} \\
        KDI & 1.24\std{0.69} & 0.82\std{0.81} &1.07\std{0.71} & 1.00\std{0.72} \\
        LDRI & 0.75\std{0.84} & 0.50\std{0.86} &0.43\std{0.87} & 0.26\std{0.28} \\
        LSRI & 0.61\std{0.70} & 0.36\std{0.66} &0.07\std{0.09} & 0.39\std{0.52} \\
        McRao & 0.63\std{0.64} & 0.38\std{0.70} &1.31\std{0.75} & 0.66\std{0.68} \\
        PBM & 0.70\std{0.72} & 0.42\std{0.59} &1.36\std{0.66} & 0.69\std{0.72} \\
        SI & 0.76\std{0.77} & \textbf{0.27\std{0.54}} &1.61\std{0.39} & 0.66\std{0.74} \\
        WGSS & 0.55\std{0.71} & 0.37\std{0.67} &\textbf{0.06\std{0.09}} & 0.39\std{0.52} \\
        WG & 1.10\std{0.66} & 0.47\std{0.57} &1.90\std{0.06} & 0.84\std{0.64} \\
        XBI & 0.73\std{0.79} & 0.35\std{0.53} &1.65\std{0.25} & 0.53\std{0.59} \\
        \cmidrule{2-5}
        H & 0.44\std{0.72} & 0.90\std{0.93} &1.50\std{0.68} & 0.29\std{0.38} \\
        KL & 0.44\std{0.71} & 0.90\std{0.92} &1.50\std{0.68} & 0.29\std{0.36} \\
        TV & 0.44\std{0.72} & 0.91\std{0.94} &1.50\std{0.68} & 0.29\std{0.40} \\
        Binary & \textbf{0.12\std{0.24}} & 0.93\std{0.99} &0.52\std{0.66} & \textbf{0.24\std{0.37}} \\
        \bottomrule
    \end{tabular}}
\end{table}

%% file: figures/table_benchmark_mixed_pearson_extended.tex
\begin{table}[th]
    \centering
    \caption{Extended results for the average Pearson correlation\std{std} ($\uparrow$) of the ranking metrics when the base clusterings seek different number of clusters.}
    \label{tab:benchmark_mixed_pearson_extended}
    \resizebox{!}{0.8\height}{
    \begin{tabular}{lllllllll}
        \toprule
        &\multicolumn{4}{c}{FCPS}&\multicolumn{4}{c}{UCI}\\
        \cmidrule(r){2-5}\cmidrule(l){6-9}
        Model & Agg. & DBSCAN & Kmeans & Alltogether& Agg. & DBSCAN & Kmeans & Alltogether \\
        \midrule
        AARI & 0.40\std{0.43} & 0.59\std{0.64} & 0.00\std{0.67} & 0.33\std{0.52} &0.20\std{0.49} & 0.21\std{0.80} & 0.13\std{0.58} & 0.43\std{0.40} \\
        ANMI & 0.26\std{0.48} & 0.61\std{0.65} & -0.08\std{0.67} & 0.20\std{0.52} &0.26\std{0.50} & 0.26\std{0.79} & -0.02\std{0.65} & 0.50\std{0.31} \\
        BHI & 0.14\std{0.41} & -0.08\std{0.96} & -0.11\std{0.59} & -0.05\std{0.53} &0.17\std{0.36} & 0.54\std{0.58} & -0.14\std{0.62} & 0.19\std{0.55} \\
        BRI & -0.40\std{0.45} & -0.18\std{0.89} & -0.41\std{0.56} & -0.22\std{0.28} &0.36\std{0.26} & 0.43\std{0.66} & 0.17\std{0.33} & 0.19\std{0.23} \\
        CHI & -0.05\std{0.55} & 0.43\std{0.80} & 0.27\std{0.49} & -0.08\std{0.62} &0.61\std{0.33} & 0.60\std{0.37} & 0.33\std{0.60} & 0.53\std{0.35} \\
        CI & 0.02\std{0.62} & -0.13\std{0.92} & -0.03\std{0.64} & -0.14\std{0.65} &-0.07\std{0.51} & 0.30\std{0.77} & 0.00\std{0.60} & 0.28\std{0.30} \\
        DBI & 0.06\std{0.57} & -0.34\std{0.78} & 0.25\std{0.45} & -0.16\std{0.50} &-0.49\std{0.44} & -0.23\std{0.42} & 0.16\std{0.42} & -0.15\std{0.36} \\
        DHI & 0.19\std{0.41} & -0.37\std{0.67} & -0.35\std{0.52} & -0.09\std{0.46} &-0.55\std{0.23} & 0.09\std{0.61} & -0.27\std{0.56} & -0.23\std{0.30} \\
        DI & 0.20\std{0.48} & 0.23\std{0.81} & 0.07\std{0.36} & 0.26\std{0.44} &-0.56\std{0.29} & -0.10\std{0.70} & 0.06\std{0.34} & -0.26\std{0.28} \\
        DRI & -0.36\std{0.29} & -0.05\std{0.94} & -0.53\std{0.28} & -0.41\std{0.23} &-0.14\std{0.39} & 0.62\std{0.39} & -0.36\std{0.34} & -0.12\std{0.29} \\
        HI & 0.21\std{0.44} & 0.30\std{0.63} & 0.57\std{0.33} & 0.24\std{0.42} &-0.33\std{0.55} & -0.50\std{0.48} & 0.32\std{0.62} & -0.16\std{0.50} \\
        KDI & 0.05\std{0.35} & 0.29\std{0.74} & 0.54\std{0.30} & 0.23\std{0.27} &-0.05\std{0.45} & -0.01\std{0.55} & 0.26\std{0.22} & 0.06\std{0.26} \\
        LDRI & -0.20\std{0.55} & -0.04\std{1.01} & -0.45\std{0.53} & -0.32\std{0.54} &0.15\std{0.66} & 0.71\std{0.48} & -0.46\std{0.46} & 0.09\std{0.48} \\
        LSRI & -0.06\std{0.59} & -0.09\std{0.96} & -0.35\std{0.55} & -0.21\std{0.60} &0.43\std{0.63} & 0.75\std{0.57} & -0.24\std{0.65} & 0.32\std{0.59} \\
        McRao & -0.13\std{0.52} & -0.26\std{0.87} & -0.26\std{0.59} & -0.26\std{0.55} &-0.16\std{0.61} & 0.27\std{0.73} & -0.19\std{0.60} & 0.08\std{0.46} \\
        PBM & 0.00\std{0.50} & 0.26\std{0.71} & 0.45\std{0.56} & 0.07\std{0.55} &-0.15\std{0.27} & -0.08\std{0.76} & 0.47\std{0.31} & -0.04\std{0.31} \\
        SI & -0.08\std{0.56} & 0.15\std{0.77} & \textbf{0.72\std{0.36}} & -0.10\std{0.63} &-0.34\std{0.38} & -0.06\std{0.69} & 0.42\std{0.45} & -0.12\std{0.35} \\
        WGSS & -0.03\std{0.60} & -0.07\std{0.99} & -0.23\std{0.58} & -0.18\std{0.62} &0.45\std{0.62} & \textbf{0.76\std{0.58}} & -0.24\std{0.64} & 0.33\std{0.57} \\
        WG & 0.38\std{0.38} & -0.33\std{0.70} & -0.32\std{0.45} & 0.10\std{0.41} &-0.34\std{0.20} & 0.08\std{0.60} & -0.23\std{0.48} & -0.23\std{0.19} \\
        XBI & -0.34\std{0.45} & 0.12\std{0.74} & 0.31\std{0.26} & -0.27\std{0.42} &-0.38\std{0.35} & -0.18\std{0.67} & 0.22\std{0.40} & -0.10\std{0.35} \\
        \cmidrule{2-9}
        H & 0.26\std{0.55} & 0.51\std{0.73} & -0.12\std{0.69} & 0.25\std{0.58} &-0.12\std{0.67} & 0.13\std{0.82} & -0.07\std{0.63} & 0.39\std{0.57} \\
        KL & 0.30\std{0.50} & 0.50\std{0.74} & -0.10\std{0.68} & 0.29\std{0.54} &-0.12\std{0.66} & 0.13\std{0.81} & -0.05\std{0.61} & 0.38\std{0.56} \\
        TV & 0.24\std{0.57} & 0.51\std{0.73} & -0.13\std{0.69} & 0.23\std{0.60} &-0.13\std{0.68} & 0.13\std{0.82} & -0.08\std{0.63} & 0.39\std{0.57} \\
        Binary & \textbf{0.80\std{0.25}} & \textbf{0.90\std{0.19}} & \textbf{0.72\std{0.21}} & \textbf{0.70\std{0.37}} &\textbf{0.67\std{0.39}} & 0.48\std{0.58} & \textbf{0.54\std{0.39}} & \textbf{0.66\std{0.25}} \\
        \bottomrule
    \end{tabular}}
\end{table}

%% file: figures/table_benchmark_mixed_regretpearson_extended.tex
\begin{table}[th]
    \centering
    \caption{Extended results for the regret on the Perason correlation\std{std} ($\downarrow$) of the ranking metrics when the base clusterings seek different number of clusters.}
    \label{tab:benchmark_mixed_regretpearson_extended}
    \resizebox{!}{0.8\height}{
    \begin{tabular}{lllllllll}
        \toprule
        &\multicolumn{4}{c}{FCPS}&\multicolumn{4}{c}{UCI}\\
        \cmidrule(r){2-5}\cmidrule(l){6-9}
        Model & Agg. & DBSCAN & Kmeans & Alltogether& Agg. & DBSCAN & Kmeans & Alltogether \\
        \midrule
        AARI & 0.46\std{0.39} & 0.37\std{0.60} & 0.89\std{0.65} & 0.47\std{0.50} &0.65\std{0.51} & 0.75\std{0.79} & 0.66\std{0.62} & 0.37\std{0.40} \\
        ANMI & 0.60\std{0.47} & 0.35\std{0.61} & 0.97\std{0.65} & 0.60\std{0.54} &0.59\std{0.51} & 0.71\std{0.78} & 0.81\std{0.69} & 0.30\std{0.20} \\
        BHI & 0.72\std{0.41} & 1.04\std{0.98} & 1.01\std{0.58} & 0.86\std{0.60} &0.67\std{0.35} & 0.43\std{0.59} & 0.92\std{0.68} & 0.61\std{0.48} \\
        BRI & 1.26\std{0.45} & 1.14\std{0.90} & 1.30\std{0.55} & 1.02\std{0.36} &0.49\std{0.32} & 0.53\std{0.65} & 0.62\std{0.41} & 0.62\std{0.31} \\
        CHI & 0.91\std{0.54} & 0.53\std{0.80} & 0.62\std{0.47} & 0.89\std{0.66} &0.24\std{0.36} & 0.37\std{0.38} & 0.45\std{0.64} & 0.27\std{0.31} \\
        CI & 0.84\std{0.58} & 1.10\std{0.94} & 0.92\std{0.62} & 0.94\std{0.71} &0.92\std{0.50} & 0.67\std{0.78} & 0.78\std{0.67} & 0.53\std{0.31} \\
        DBI & 0.81\std{0.48} & 1.30\std{0.77} & 0.64\std{0.42} & 0.97\std{0.55} &1.34\std{0.41} & 1.20\std{0.44} & 0.63\std{0.36} & 0.96\std{0.32} \\
        DHI & 0.68\std{0.42} & 1.33\std{0.67} & 1.24\std{0.51} & 0.89\std{0.58} &1.40\std{0.24} & 0.88\std{0.62} & 1.06\std{0.64} & 1.03\std{0.28} \\
        DI & 0.66\std{0.52} & 0.73\std{0.78} & 0.83\std{0.33} & 0.55\std{0.43} &1.41\std{0.27} & 1.07\std{0.71} & 0.73\std{0.43} & 1.07\std{0.38} \\
        DRI & 1.22\std{0.36} & 1.01\std{0.95} & 1.43\std{0.27} & 1.21\std{0.31} &1.04\std{0.43} & 0.34\std{0.35} & 1.10\std{0.65} & 0.93\std{0.33} \\
        HI & 0.65\std{0.40} & 0.66\std{0.61} & 0.32\std{0.34} & 0.56\std{0.47} &1.17\std{0.55} & 1.47\std{0.51} & 0.46\std{0.59} & 0.97\std{0.57} \\
        KDI & 0.81\std{0.37} & 0.68\std{0.72} & 0.36\std{0.32} & 0.58\std{0.23} &0.94\std{0.41} & 0.99\std{0.54} & 0.60\std{0.25} & 0.78\std{0.30} \\
        LDRI & 1.07\std{0.56} & 1.00\std{1.03} & 1.34\std{0.52} & 1.13\std{0.56} &0.75\std{0.71} & 0.24\std{0.43} & 1.20\std{0.76} & 0.72\std{0.47} \\
        LSRI & 0.92\std{0.58} & 1.06\std{0.97} & 1.25\std{0.54} & 1.01\std{0.65} &0.42\std{0.60} & 0.22\std{0.58} & 1.03\std{0.72} & 0.49\std{0.52} \\
        McRao & 0.99\std{0.50} & 1.23\std{0.87} & 1.16\std{0.57} & 1.06\std{0.60} &1.01\std{0.62} & 0.70\std{0.73} & 0.97\std{0.69} & 0.73\std{0.48} \\
        PBM & 0.86\std{0.46} & 0.70\std{0.69} & 0.45\std{0.55} & 0.74\std{0.64} &1.00\std{0.27} & 1.05\std{0.78} & 0.32\std{0.17} & 0.85\std{0.32} \\
        SI & 0.95\std{0.52} & 0.82\std{0.75} & \textbf{0.18\std{0.33}} & 0.91\std{0.65} &1.19\std{0.34} & 1.02\std{0.71} & 0.36\std{0.33} & 0.92\std{0.31} \\
        WGSS & 0.89\std{0.58} & 1.03\std{1.01} & 1.13\std{0.56} & 0.99\std{0.66} &0.40\std{0.59} & \textbf{0.20\std{0.59}} & 1.02\std{0.71} & 0.48\std{0.49} \\
        WG & 0.48\std{0.35} & 1.30\std{0.71} & 1.21\std{0.43} & 0.71\std{0.49} &1.19\std{0.21} & 0.88\std{0.59} & 1.02\std{0.54} & 1.03\std{0.20} \\
        XBI & 1.20\std{0.45} & 0.84\std{0.72} & 0.58\std{0.25} & 1.07\std{0.45} &1.23\std{0.31} & 1.14\std{0.69} & 0.56\std{0.36} & 0.91\std{0.32} \\
        \cmidrule{2-9}
        H & 0.60\std{0.55} & 0.45\std{0.69} & 1.02\std{0.67} & 0.56\std{0.63} &0.97\std{0.71} & 0.83\std{0.81} & 0.86\std{0.68} & 0.42\std{0.59} \\
        KL & 0.56\std{0.50} & 0.46\std{0.70} & 0.99\std{0.66} & 0.52\std{0.60} &0.96\std{0.70} & 0.84\std{0.81} & 0.84\std{0.67} & 0.42\std{0.59} \\
        TV & 0.62\std{0.57} & 0.45\std{0.69} & 1.03\std{0.68} & 0.58\std{0.65} &0.97\std{0.72} & 0.83\std{0.81} & 0.87\std{0.68} & 0.42\std{0.59} \\
        Binary & \textbf{0.06\std{0.16}} & \textbf{0.07\std{0.16}} & \textbf{0.18\std{0.22}} & \textbf{0.11\std{0.24}} &\textbf{0.18\std{0.38}} & 0.49\std{0.57} & \textbf{0.25\std{0.24}} & \textbf{0.14\std{0.16}} \\
        \bottomrule
    \end{tabular}}
\end{table}

%% file: figures/table_benchmark_mixed_regretari_extended.tex
\begin{table}[th]
    \centering
    \caption{Extended results for the regret on the ARI of the model selected\std{std} ($\downarrow$) by each ranking metric when the base clusterings seek different number of clusters.}
    \label{tab:benchmark_mixed_regretari_extended}
    \resizebox{!}{0.8\height}{
    \begin{tabular}{lllllllll}
        \toprule
        &\multicolumn{4}{c}{FCPS}&\multicolumn{4}{c}{UCI}\\
        \cmidrule(r){2-5}\cmidrule(l){6-9}
        Model & Agg. & DBSCAN & Kmeans & Alltogether& Agg. & DBSCAN & Kmeans & Alltogether \\
        \midrule
            AARI & 0.41\std{0.32} & 0.10\std{0.17} & 0.33\std{0.27} & 0.41\std{0.32} &0.20\std{0.24} & 0.11\std{0.13} & 0.20\std{0.19} & 0.20\std{0.25} \\
            ANMI & 0.41\std{0.33} & 0.10\std{0.17} & 0.33\std{0.30} & 0.40\std{0.34} &0.17\std{0.23} & 0.12\std{0.13} & 0.19\std{0.24} & 0.14\std{0.15} \\
            BHI & 0.66\std{0.21} & 0.25\std{0.21} & 0.54\std{0.26} & 0.69\std{0.20} &0.23\std{0.22} & 0.14\std{0.23} & 0.30\std{0.26} & 0.26\std{0.24} \\
            BRI & 0.67\std{0.21} & 0.24\std{0.18} & 0.58\std{0.20} & 0.72\std{0.18} &0.21\std{0.21} & 0.14\std{0.19} & 0.29\std{0.27} & 0.26\std{0.24} \\
            CHI & 0.24\std{0.32} & 0.07\std{0.15} & 0.16\std{0.19} & 0.27\std{0.34} &0.13\std{0.20} & 0.06\std{0.09} & 0.15\std{0.19} & 0.12\std{0.17} \\
            CI & 0.36\std{0.37} & 0.20\std{0.20} & 0.23\std{0.28} & 0.38\std{0.36} &0.30\std{0.23} & 0.14\std{0.15} & 0.25\std{0.22} & 0.32\std{0.22} \\
            DBI & 0.30\std{0.33} & 0.29\std{0.33} & 0.19\std{0.19} & 0.31\std{0.35} &0.43\std{0.27} & 0.24\std{0.24} & 0.16\std{0.24} & 0.45\std{0.29} \\
            DHI & 0.25\std{0.25} & 0.33\std{0.27} & 0.54\std{0.26} & 0.26\std{0.28} &0.39\std{0.26} & 0.22\std{0.23} & 0.27\std{0.24} & 0.41\std{0.28} \\
            DI & 0.24\std{0.37} & 0.17\std{0.35} & 0.29\std{0.25} & 0.25\std{0.38} &0.43\std{0.28} & 0.20\std{0.16} & 0.26\std{0.30} & 0.45\std{0.29} \\
            DRI & 0.67\std{0.21} & 0.20\std{0.20} & 0.54\std{0.25} & 0.69\std{0.20} &0.36\std{0.20} & 0.13\std{0.24} & 0.37\std{0.25} & 0.38\std{0.23} \\
            HI & 0.62\std{0.39} & 0.20\std{0.34} & 0.16\std{0.25} & 0.58\std{0.37} &0.43\std{0.27} & 0.28\std{0.22} & 0.27\std{0.27} & 0.45\std{0.29} \\
            KDI & 0.56\std{0.35} & 0.19\std{0.34} & 0.44\std{0.31} & 0.59\std{0.35} &0.21\std{0.25} & 0.13\std{0.13} & 0.21\std{0.27} & 0.29\std{0.26} \\
            LDRI & 0.67\std{0.21} & 0.20\std{0.20} & 0.54\std{0.25} & 0.69\std{0.20} &0.36\std{0.20} & 0.13\std{0.24} & 0.37\std{0.25} & 0.38\std{0.23} \\
            LSRI & 0.66\std{0.21} & 0.20\std{0.20} & 0.54\std{0.26} & 0.69\std{0.20} &0.23\std{0.22} & 0.07\std{0.19} & 0.30\std{0.26} & 0.26\std{0.24} \\
            McRao & 0.66\std{0.21} & 0.21\std{0.19} & 0.55\std{0.25} & 0.69\std{0.19} &0.37\std{0.23} & 0.08\std{0.11} & 0.30\std{0.26} & 0.40\std{0.23} \\
            PBM & 0.26\std{0.32} & 0.21\std{0.34} & 0.16\std{0.17} & 0.27\std{0.34} &0.41\std{0.26} & 0.18\std{0.17} & 0.21\std{0.27} & 0.42\std{0.28} \\
            SI & 0.24\std{0.34} & 0.08\std{0.17} & \textbf{0.10\std{0.14}} & 0.31\std{0.34} &0.40\std{0.26} & 0.19\std{0.19} & 0.14\std{0.24} & 0.41\std{0.28} \\
            WGSS & 0.66\std{0.21} & 0.20\std{0.20} & 0.54\std{0.26} & 0.69\std{0.20} &0.23\std{0.22} & 0.07\std{0.19} & 0.30\std{0.26} & 0.26\std{0.24} \\
            WG & 0.21\std{0.23} & 0.29\std{0.19} & 0.54\std{0.26} & \textbf{0.22\std{0.21}} &0.40\std{0.26} & 0.18\std{0.23} & 0.28\std{0.25} & 0.41\std{0.28} \\
            XBI & 0.28\std{0.31} & 0.19\std{0.33} & 0.15\std{0.19} & 0.34\std{0.38} &0.40\std{0.26} & 0.21\std{0.24} & 0.16\std{0.23} & 0.41\std{0.28} \\
        \cmidrule{2-9}
            H & 0.40\std{0.32} & 0.10\std{0.17} & 0.41\std{0.27} & 0.43\std{0.31} &0.23\std{0.25} & 0.12\std{0.13} & 0.23\std{0.23} & 0.15\std{0.21} \\
            KL & 0.40\std{0.32} & 0.10\std{0.17} & 0.42\std{0.28} & 0.43\std{0.31} &0.23\std{0.25} & 0.12\std{0.13} & 0.23\std{0.23} & 0.15\std{0.21} \\
            TV & 0.40\std{0.32} & 0.10\std{0.17} & 0.41\std{0.27} & 0.43\std{0.31} &0.23\std{0.25} & 0.12\std{0.13} & 0.23\std{0.23} & 0.15\std{0.21} \\
            Binary & \textbf{0.14\std{0.20}} & \textbf{0.06\std{0.14}} & 0.17\std{0.21} & \textbf{0.22\std{0.24}} &\textbf{0.11\std{0.16}} & \textbf{0.05\std{0.06}} & \textbf{0.09\std{0.12}} & \textbf{0.10\std{0.14}} \\
        \bottomrule
    \end{tabular}}
\end{table}

%% file: figures/table_benchmark_fixed_regretari_extended.tex
\begin{table}[th]
    \centering
    \caption{Extended results for the regret on the ARI of the model selected\std{std} ($\downarrow$) by each ranking metric when the base clusterings are restricted to as many clusters as targets.}
    \label{tab:benchmark_fixed_regretari_extended}
    \resizebox{!}{0.8\height}{
    \begin{tabular}{lllll}
        \toprule
        &\multicolumn{2}{c}{FCPS}&\multicolumn{2}{c}{UCI}\\
        \cmidrule(r){2-3}\cmidrule(l){4-5}
        Model & Agglomerative & Kmeans & Agglomerative & Kmeans \\
        \midrule
        AARI & 0.16\std{0.28} & 0.06\std{0.10} &0.31\std{0.22} & 0.05\std{0.05} \\
        ANMI & 0.16\std{0.28} & 0.06\std{0.10} &0.31\std{0.22} & 0.03\std{0.03} \\
        BHI & 0.24\std{0.33} & 0.05\std{0.09} &\textbf{0.00\std{0.00}} & 0.06\std{0.05} \\
        BRI & \textbf{0.14\std{0.21}} & 0.07\std{0.12} &\textbf{0.00\std{0.00}} & \textbf{0.02\std{0.03}} \\
        CHI & 0.24\std{0.33} & 0.05\std{0.09} &\textbf{0.00\std{0.00}} & 0.06\std{0.05} \\
        CI & 0.22\std{0.27} & 0.05\std{0.09} &0.29\std{0.23} & 0.11\std{0.11} \\
        DBI & 0.36\std{0.41} & 0.02\std{0.04} &0.43\std{0.28} & 0.09\std{0.08} \\
        DHI & 0.48\std{0.43} & 0.02\std{0.04} &0.43\std{0.28} & 0.18\std{0.16} \\
        DI & 0.09\std{0.27} & \textbf{0.01\std{0.01}} &0.38\std{0.25} & 0.08\std{0.06} \\
        DRI & 0.30\std{0.41} & 0.05\std{0.10} &0.23\std{0.27} & 0.05\std{0.07} \\
        HI & 0.36\std{0.41} & 0.02\std{0.04} &0.43\std{0.28} & 0.15\std{0.16} \\
        KDI & 0.57\std{0.40} & 0.14\std{0.25} &0.17\std{0.18} & 0.17\std{0.18} \\
        LDRI & 0.30\std{0.41} & 0.05\std{0.10} &0.23\std{0.27} & 0.05\std{0.07} \\
        LSRI & 0.24\std{0.33} & 0.05\std{0.09} &\textbf{0.00\std{0.00}} & 0.06\std{0.05} \\
        McRao & 0.30\std{0.32} & 0.05\std{0.09} &0.28\std{0.25} & 0.09\std{0.06} \\
        PBM & 0.37\std{0.38} & 0.05\std{0.09} &0.35\std{0.27} & 0.12\std{0.17} \\
        SI & 0.31\std{0.37} & 0.02\std{0.04} &0.38\std{0.25} & 0.07\std{0.08} \\
        WGSS & 0.24\std{0.33} & 0.05\std{0.09} &\textbf{0.00\std{0.00}} & 0.06\std{0.05} \\
        WG & 0.57\std{0.45} & 0.01\std{0.04} &0.43\std{0.28} & 0.16\std{0.16} \\
        XBI & 0.31\std{0.37} & 0.02\std{0.04} &0.42\std{0.28} & 0.06\std{0.07} \\
        \cmidrule{2-5}
        H & 0.26\std{0.37} & 0.06\std{0.10} &0.32\std{0.21} & 0.05\std{0.05} \\
        KL & 0.26\std{0.37} & 0.06\std{0.10} &0.32\std{0.21} & 0.05\std{0.05} \\
        TV & 0.26\std{0.37} & 0.06\std{0.10} &0.32\std{0.21} & 0.05\std{0.05} \\
        Binary & 0.15\std{0.24} & 0.05\std{0.10} &0.17\std{0.18} & 0.05\std{0.06} \\
        \bottomrule
    \end{tabular}}
\end{table}